\documentclass{dukecei}
\usepackage[utf8]{inputenc}
\usepackage{url}
\usepackage{microtype}

\renewcommand{\eqref}[1]{equation~\ref{#1}}

\title{BasisKV: Value Latents for Communication and Key Routing}
\author{%
Lixun Zhang$^{1}$,
Zishan Shao$^{1,\dagger}$,
Kangning Cui$^{2}$,
Yu Liang$^{1}$,
Fan Yang$^{2}$,
Yiran Chen$^{1}$
\\[0.65em]
\normalfont\small
$^{1}$Duke University
\quad
$^{2}$Wake Forest University
}
\date{}

\begin{document}

\maketitle
\footnotetext[2]{Corresponding Email: zishan.shao@duke.edu}
\thispagestyle{firstpagestyle}

\begin{abstract}

Long-context large language model (LLM) inference is increasingly constrained by the cost of maintaining an ever-growing KV-cache. Low-rank K/V projections provide a promising way to compress the KV-cache with minimal quality loss, but prior work has largely optimized KV-cache compression in isolation, leaving distributed inference, especially with tensor parallelism where attention computation is partitioned across multiple GPUs, largely underexplored. We make two observations: (i) tensor-parallel communication is an important cost when deploying low-rank KV-cache compression in distributed inference; and (ii) Value compression can be associated with the output projection, allowing the same low-rank Value representation to reduce both Value-cache width and tensor-parallel communication. Based on these observations, we propose \textbf{BasisKV}, a low-rank KV-cache compression method for communication-intensive, long-context LLM inference. BasisKV jointly optimizes the low-rank factors across per-GPU
attention partitions against the complete attention output, rather than fitting each partition in isolation. The compressed Value cache is further reused to help identify which Key pages should be accessed: it predicts part of the Key information needed for routing, while a Key-specific residual stores what cannot be recovered from Value. The original Keys are then retrieved only for the selected pages to compute exact QK scores. Across multiple LLM families and system settings, BasisKV achieves a better quality--efficiency tradeoff than existing low-rank baselines, reaching up to \(1.42\times\) end-to-end speedup under 8-GPU tensor-parallel inference. The code is available at \url{https://github.com/alexz949/BasisKV}
\end{abstract}

\section{Introduction}

Long-context large language model (LLM) inference is increasingly constrained by the cost of maintaining an ever-growing KV cache ~\citep{liu2024kivi,chang2025palu}. A promising approach is to factorize or transform the Key and Value
projection matrices such that inference directly produces and caches
low-dimensional K/V representations
\citep{chang2025palu,saxena2024eigen}.
This broader family of low-rank KV-cache methods has attracted growing
interest, including trainable feature-space projections
\citep{lin2025matryoshkakv} and cross-layer low-rank cache representations
\citep{chang2026xkv}. However, prior low-rank KV-cache methods largely optimize cache efficiency in isolation from tensor-parallel communication. Under tensor parallelism (TP), attention computation is partitioned across multiple GPUs, and the resulting partial attention outputs must additionally be communicated across accelerators ~\citep{shoeybi2019megatron,hansenpalmus2024communication}. 
This raises a natural question: \textbf{\emph{can low-rank KV-cache compression provide additional benefits under distributed inference beyond reducing the cache itself?}}
Our key observation is that low-rank Value compression can also reduce the width of the attention outputs communicated across GPUs, allowing the same latent representation to reduce both KV-cache size and tensor-parallel traffic. This motivates \textbf{BasisKV}, a low-rank KV-cache compression method designed for distributed long-context inference.

BasisKV first moves tensor-parallel communication into a low-rank space by projecting each per-GPU attention output (\emph{source}) to a lower dimension before communication, preserving these source-specific latents through a source-preserving AllGather, and reconstructing them afterward. Rather than fitting each source independently, BasisKV jointly optimizes
all low-rank factors against the complete attention output, accounting for the cross-source interactions that contribute to the
reconstruction error.

For attention blocks where the Value path remains linear, the low-rank compression projection applied before communication (the \emph{encoder}) can be moved to the Value side and folded into the Value projection. This allows the model to directly produce and cache the same compressed Values used for TP communication, with no additional approximation beyond the offline derived low-rank model. We further reuse the compressed Value cache for Key routing. Compressed Values provide part of the information needed to select
Key pages, while a Key-specific residual captures the remaining
routing information. Original Keys are retrieved only for selected pages to recompute exact QK scores.

Across multiple LLM families, model scales, and inference settings, BasisKV consistently preserves stronger downstream quality than matched low-rank baselines while simultaneously reducing tensor-parallel communication. On eight NVIDIA L40S GPUs, BasisKV reaches up to $1.42\times$ end-to-end speedup under tensor-parallel inference. For long-context decoding, the Value-assisted Key routing avoids accessing full Keys over most of the context; at 128K context on a single L40S GPU, BasisKV achieves $1.63\times$ attention-block speedup and $1.25\times$ full-model decode speedup.

Our contributions are threefold. First, \textbf{\emph{to our knowledge, BasisKV is the first post-training
method to jointly reduce persistent Value-cache width and tensor-parallel
attention-output communication using the same low-rank Value representation
for standard attention.}} Second, we connect low-rank Value compression with tensor-parallel communication and jointly optimize the resulting low-rank factors against the complete attention output. Third, we further reuse the compressed Value cache for sparse Key routing and demonstrate improved quality--efficiency tradeoffs and end-to-end speedups across multiple models and inference settings.

\begin{figure}[t]
    \centering
    \setlength{\abovecaptionskip}{0pt}
    \includegraphics[width=\linewidth]{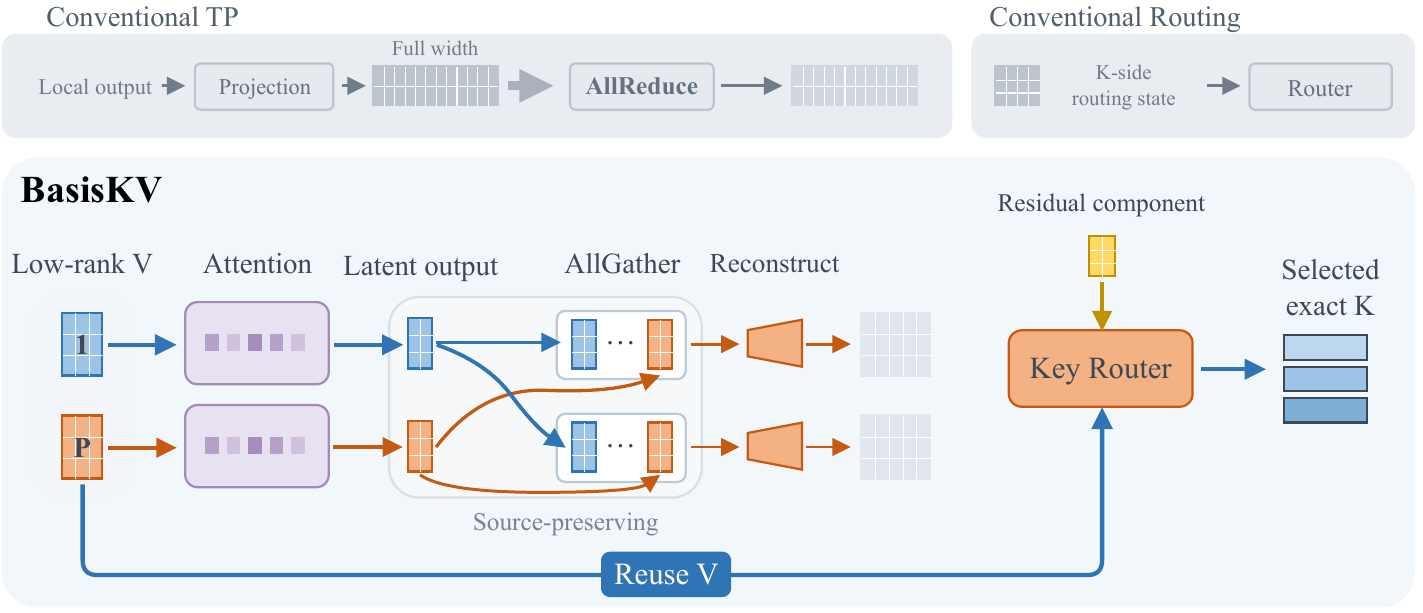}
    \caption{
\textbf{Overview of BasisKV.}
BasisKV compresses each TP-local attention output into a low-rank Value latent, communicates the source-specific latents with AllGather, and
reconstructs the complete output after communication.
The same Value latent is retained in the cache and reused for Key routing where a Key-specific residual provides the remaining
routing information.}
    \label{fig:overview}
\end{figure}

\section{Background}
\subsection{Tensor-Parallel Attention Output Projection}
\label{sec:tp_attention_output}

We consider standard tensor parallelism (TP)~\citep{hansenpalmus2024communication}
over $P$ GPUs.
Let $T$ denote the sequence length, $d$ the model width, $H_q$ the number
of query heads, and $d_h$ the head dimension.
We denote the concatenated attention-output width by
$d_a = H_q d_h$.
For both Multi-Head Attention (MHA) and Grouped-Query Attention (GQA), let
$Z^{(h)} \in \mathbb{R}^{T\times d_h}$ denote the output of query head $h$.
The concatenated attention output before the output projection is
\begin{equation*}
    Z
    =
    \operatorname{Concat}(Z^{(1)},\ldots,Z^{(H_q)})
    \in
    \mathbb{R}^{T\times d_a}.
\end{equation*}
Under TP, the QKV projections are column-parallel, so the
query heads are partitioned across GPUs.
Let $Z_p$ denote the concatenated outputs of the query heads assigned to
GPU $p$.
The output projection $W_O$ is partitioned along its input dimension to
match this sharding.
With
$Z_p\in\mathbb{R}^{T\times(d_a/P)}$
and
$W_{O,p}\in\mathbb{R}^{(d_a/P)\times d}$,
the corresponding partitions are
\begin{equation*}
    Z
    =
    \begin{bmatrix}
        Z_1 & \cdots & Z_P
    \end{bmatrix},
    \qquad
    W_O
    =
    \begin{bmatrix}
        W_{O,1}\\
        \vdots\\
        W_{O,P}
    \end{bmatrix}.
\end{equation*}
Each GPU holds only a $d_a/P$-dimensional shard of the attention
output, but its row-parallel block $W_{O,p}$ still maps this local shard to a full $d$-dimensional contribution,
\begin{equation*}
    Y_p = Z_p W_{O,p}
    \in \mathbb{R}^{T\times d}.
\end{equation*}
The partial contributions are then summed across GPUs using AllReduce,
yielding the complete attention output on every GPU:
\begin{equation*}
    Y
    =
    \operatorname{AllReduce}(Y_p)
    =
    \sum_{p=1}^{P} Y_p
    =
    \sum_{p=1}^{P} Z_p W_{O,p}.
    \label{eq:dense_allreduce}
\end{equation*}
Thus, although the attention output remains sharded at width $d_a/P$
before $W_O$, the standard row-parallel execution projects each local
shard to a full-width $d$-dimensional contribution before communication.
Our goal is to move the communication boundary earlier by communicating a
lower-dimensional local representation and reconstructing the full-width
output only after the collective.

\subsection{Low-Rank TP Communication}
\label{sec:low_rank_comm}

Consider a rank-$r$ factorization of the local output-projection matrix,
\begin{equation*}
    W_{O,p}
    \approx
    \widehat{W}_{O,p}
    =
    E_pD_p,
    \qquad
    E_p\in\mathbb{R}^{(d_a/P)\times r},
    \qquad
    D_p\in\mathbb{R}^{r\times d}.
\end{equation*}
If both factors are evaluated locally, the partial output
$Z_pE_pD_p\in\mathbb{R}^{T\times d}$ remains full width.
Thus, low-rank factorization of $W_{O,p}$ alone does not reduce TP
communication.

To reduce communication, we place the collective between the two factors.
GPU $p$ first forms
$
    U_p
    =
    Z_pE_p
    \in
    \mathbb{R}^{T\times r}.
$
If $U_p$ is communicated using AllReduce, the TP-source representations
are summed before reconstruction:
\begin{equation*}
    \widehat{Y}_{\mathrm{AR}}
    =
    \operatorname{AllReduce}(U_p)D
    =
    \left(
        \sum_{p=1}^{P} Z_pE_p
    \right)D,
    \label{eq:lr_allreduce_rep}
\end{equation*}
where the shared decoder $D\in\mathbb{R}^{r\times d}$ reconstructs the
reduced representation.
If AllGather is used instead, the TP-source representations remain
separate through communication:
\begin{equation*}
    \widehat{Y}_{\mathrm{AG}}
    =
    \operatorname{AllGather}(U_p)
    \begin{bmatrix}
        D_1\\
        \vdots\\
        D_P
    \end{bmatrix}
    =
    \sum_{p=1}^{P} Z_pE_pD_p.
    \label{eq:lr_allgather_rep}
\end{equation*}
The individual representations remain available after communication,
allowing a separate decoder $D_p$ for each TP source.

\section{Collective-Aware Low-Rank Attention}
\label{sec:method}

Building on the Low-Rank AllGather structure in
Section~\ref{sec:low_rank_comm}, we jointly fit the TP-local low-rank
factors against the complete attention output.
For linear Value paths, the compression factors can be folded into the
Value projections, so the same low-dimensional representation reduces both
TP communication and Value-cache width.
We further allocate ranks across layers under a fixed global budget using
forward-only Kullback--Leibler (KL) sensitivity profiling.

\subsection{Joint Fitting}
\label{sec:joint_fitting}

Although each TP-local block can be approximated separately, the complete
attention output is the sum of all source contributions.
Let
$
\epsilon_p
=
Z_pW_{O,p}-Z_pE_pD_p
$
denote the approximation error from source $p$.
Then
\begin{equation*}
    \left\|Y-\widehat Y\right\|_F^2
    =
    \sum_{p=1}^{P}\|\epsilon_p\|_F^2
    +
    2\sum_{p<q}\langle\epsilon_p,\epsilon_q\rangle_F.
\end{equation*}
Independent fitting does not account for the cross-source terms that appear in the reconstruction error of the complete TP output.
We therefore optimize the complete output directly.
\begin{equation}
    \min_{\{E_p,D_p\}_{p=1}^{P}}
    \left\|
        Y-\sum_{p=1}^{P}Z_pE_pD_p
    \right\|_F^2
    \label{eq:joint_objective}
\end{equation}
where $E_p$ is the compression factor for source $p$, and $D_p$ maps the
resulting latent representation to its full-width output contribution.
The objective in~\eqref{eq:joint_objective} is bilinear in
$\{E_p\}$ and $\{D_p\}$, but becomes a linear least-squares problem in
either factor family when the other is fixed.
We therefore optimize it by alternating least squares.

\paragraph{$D$ update.}
With $\{E_p\}$ fixed, we denote
$
U=[Z_1E_1\ \cdots\ Z_PE_P]
$
and let $D$ be the corresponding stacked matrix of $D_p$ factors.
Then \eqref{eq:joint_objective} becomes
\begin{equation*}
    D^\star
    =
    \arg\min_D
    \left\|Y-UD\right\|_F^2.
\end{equation*}
We solve this least-squares problem through the normal equations
$
U^\top U D^\star=U^\top Y
$.
The coupled system is assembled from calibration covariance statistics and
solved jointly.

\paragraph{$E$ update.}
With $\{D_p\}$ fixed, we update one $E_p$ at a time.
We first define
$
R_p = Y-\sum_{q\ne p}Z_qE_qD_q
$
as the residual excluding source $p$.
With the remaining factors fixed,
\eqref{eq:joint_objective} becomes
\begin{equation*}
    E_p^\star
    =
    \arg\min_{E_p}
    \left\|
        R_p-Z_pE_pD_p
    \right\|_F^2.
\end{equation*}
The corresponding normal equation is
$
(Z_p^\top Z_p)E_p^\star(D_pD_p^\top)
=
Z_p^\top R_pD_p^\top
$.
Both updates are therefore convex linear least-squares subproblems of
\eqref{eq:joint_objective}.
In practice, the $D$ system is assembled and solved by Cholesky
factorization.
We solve the $E_p$ system by matrix-free conjugate gradient, avoiding
explicit formation of the large Kronecker-structured system.
The complete derivation and implementation details are given in
Appendix~\ref{app:als}.

\subsection{Value Compression}
\label{sec:value_compression}

The compression factor learned from the complete attention-output objective
can also be applied directly to the Value path.
Let $S_h$ and $V_h$ denote the attention matrix and Value state for head
$h$, respectively, so that
$
Z_h=S_hV_h
$.
Since $S_h$ is independent of $V_h$, the fitted approximation satisfies
\begin{equation*}
    \widehat Y
    =
    \sum_{h=1}^{H_q} S_hV_hE_hD_h
    =
    \sum_{h=1}^{H_q} S_h(V_hE_h)D_h.
\end{equation*}
Hence the compression factor $E_h$ may be applied to the Value state before
attention without changing the fitted approximation.
Defining $\widehat V_h=V_hE_h$, the compressed Value can therefore be cached directly.
For GQA, all query heads sharing the same physical KV head also share
the same Value encoder $E_g$, while retaining head-specific decoders
$D_h$. This ensures that only one compressed Value latent is stored
per physical KV head.
Since $V_h=XW_{V,h}$, we have $\widehat V_h=XW_{V,h}E_h$.
Thus $E_h$ can be folded into the Value projection by setting
$
\widehat W_{V,h}=W_{V,h}E_h
$.
The same latent representation therefore determines both the communicated
width and the cached Value width, without introducing any approximation
beyond the fitted low-rank output model.

\subsection{Rank Allocation}
\label{sec:rank_allocation}

A uniform rank assignment ignores that layers have different
sensitivity to compression.
We estimate this sensitivity around a reference rank $r_0$ using one
lower-rank and one higher-rank probe for each layer.
Let $e_\ell(r)$ denote the reconstruction error of layer $\ell$ at rank
$r$, and let $\Delta\mathcal{K}_\ell^\pm$ denote the change in output KL
relative to the reference model when only that layer is changed from
$r_0$ to $r^\pm$.
We define the two one-sided sensitivities as
\begin{equation*}
    s_\ell^\pm
    =
    \max\left(
        0,\,
        \frac{
            \Delta\mathcal{K}_\ell^\pm
        }{
            e_\ell(r^\pm)-e_\ell(r_0)
        }
    \right).
    \label{eq:two_sided_kl}
\end{equation*}
The profiling requires only forward evaluation and two single-layer probes
per layer, so its cost grows linearly with model depth.
For each candidate rank, we weight its reconstruction-error change by
$s_\ell^-$ below $r_0$ and by $s_\ell^+$ above $r_0$ to obtain a layerwise
rank cost.
We then use dynamic programming to find the minimum-cost discrete rank
assignment under the fixed global budget.
The derivation and dynamic-programming details are given in
Appendix~\ref{app:rank_allocation}.

\section{Key Routing with Value Latents}
\label{sec:key_routing}
We next examine whether the compressed Value representation can also be reused for Key routing.
We find that the compressed Values can predict a nontrivial component of the corresponding pre-RoPE Keys through a low-rank representation.
We then use this Value-derived prediction as part of the routing signal and store a low-rank residual for the remaining Key-specific information.
When compressed Values are already retained for attention, this reuse reduces the amount of dedicated per-token routing state.
The resulting approximate scores are used only to select KV-cache pages and the original Keys are then accessed to recompute QK scores on the selected pages.

\subsection{Key Prediction from Value}
\label{sec:key_prediction}

Let $\widehat V_c$ and $K_c$ denote the centered compressed Value latent
and pre-RoPE Key.
We fit the rank-$r_b$ reduced-rank regression
\begin{equation}
    M^\star
    =
    \arg\min_{\operatorname{rank}(M)\le r_b}
    \|K_c-\widehat V_cM\|_F^2.
    \label{eq:value_key_rrr}
\end{equation}
With
$
G=\widehat V_c^\top\widehat V_c
$
and
$
H=\widehat V_c^\top K_c
$,
let
$
U\Sigma V^\top
$
be the rank-$r_b$ truncated SVD of $G^{\dagger/2}H$.
Then
\begin{equation*}
    M^\star
    =
    G^{\dagger/2}U\Sigma V^\top
\end{equation*}
Therefore the predicted Key is $ \widehat K = \mu_K+\widehat V_cM^\star$.
The model's native RoPE transformation is applied after prediction.
\paragraph{Routing-aligned Value basis.}
Evaluating the predictor directly would require reading the full compressed
Value state for every cached token.
However, since $ M^\star=G^{\dagger/2}U\Sigma V^\top$,
the prediction depends on $\widehat V_c$ only through the
$r_b$-dimensional quantity $ \widehat V_cG^{\dagger/2}U\Sigma$.
We therefore align this predictive subspace with explicit coordinates of
the Value cache.
We compute the thin QR factorization
\begin{equation*}
    G^{\dagger/2}U\Sigma=Q_bR_b,
    \qquad
    M^\star=Q_bR_bV^\top.
\end{equation*}
Completing $Q_b$ to an orthogonal matrix
$
Q=[Q_b\;Q_\perp]
$,
we rotate the Value latent and decoder as
$
\overline V=\widehat VQ
$
and
$
\overline D=Q^\top D
$.
This change of basis leaves the Value reconstruction unchanged and places
all information required by the Key predictor in the first $r_b$
coordinates of $\overline V$.
The rotation is folded into the Value encoder offline, so full-context
routing reads only these $r_b$ coordinates.
The derivation is given in Appendix~\ref{app:key_routing}.

\subsection{Residual Routing Component}
\label{sec:query_residual}

After applying RoPE, let $k_t$ and $\widehat k_t$ denote the exact and
predicted Keys for token $t$, and define the residual
$
e_t=k_t-\widehat k_t.
$
The Value-based predictor does not recover the full Key, so we represent
the remaining score contribution with a low-rank residual component.
We factor this contribution into a query-side factor $U_h$ and a
residual-side factor $F_h$,
\begin{equation}
    \widehat s_{ht}
    =
    \frac{1}{\sqrt{d_h}}
    \left[
        q_h\widehat k_t^\top
        +
        (q_hU_h)(e_tF_h)^\top
    \right].
    \label{eq:query_residual_score}
\end{equation}
Thus, only the $r_q$-dimensional residual representation $e_tF_h$ is
stored for each cached token.

\paragraph{Page-Fisher fitting.}
The residual correction is used to select pages rather than to reconstruct
the Key itself.
Let $\ell$ and $\widehat\ell$ denote the exact and approximate page logits.
We measure the routing error by
\begin{equation*}
    D_{\mathrm{KL}}
    \left(
        \operatorname{softmax}(\ell)
        \,\middle\|\,
        \operatorname{softmax}(\widehat\ell)
    \right),
\end{equation*}
which measures how the approximation redistributes attention mass among
pages.
Direct optimization of this objective is nonlinear in the residual
factors.
However, a second-order expansion around the exact page logits gives a
quadratic objective for the residual correction.

Fix one calibration query and head.
Let $\pi_p$ denote the exact attention mass of page $p$ and
$\bar e_p$ be the attention-weighted average Key residual within page $p$.
Define
\begin{equation*}
    \bar e
    =
    \sum_p \pi_p\bar e_p,
    \qquad
    \Gamma
    =
    \sum_p
    \pi_p
    (\bar e_p-\bar e)^\top
    (\bar e_p-\bar e).
\end{equation*}
Here $\bar e$ is the mean residual over pages, and $\Gamma$ is the
corresponding covariance.
The second-order expansion gives
\begin{equation}
    \mathcal L_{\mathrm{PF}}
    =
    \frac{1}{2d_h}
    q_h
    (U_hF_h^\top-I)
    \Gamma
    (U_hF_h^\top-I)^\top
    q_h^\top,
    \label{eq:page_fisher}
\end{equation}
$I$ is the identity matrix corresponding to the exact residual score, while $U_hF_h^\top$ gives its low-rank approximation.
We sum this loss over calibration queries and heads.
For fixed $F_h$, \eqref{eq:page_fisher} is a convex quadratic problem in
$U_h$, and vice versa.
We alternate between the two factors and solve the resulting
positive-semidefinite systems by matrix-free conjugate gradient.
The KL expansion and normal equations are given in
Appendix~\ref{app:key_routing}.
\paragraph{Query selection.}
Fitting the residual at every query position is expensive and provides
many redundant observations.
We therefore select a small set of representative positions.

For each query head $h$, we first compute
$
H_h=\mathbb E[q_h^\top q_h]
$
over the calibration queries.
The factor $H_h^\dagger$ balances query directions with different scales
when comparing positions.
Let $Q_h$ be the stacked matrix of candidate queries for head $h$.
We then define the position Gram matrix
\begin{equation}
    G
    =
    \frac{1}{H_q}
    \sum_{h=1}^{H_q}
    \mathbb E
    \left[
        Q_hH_h^\dagger Q_h^\top
    \right].
    \label{eq:query_gram}
\end{equation}
Thus, $G$ measures similarity between candidate positions after accounting
for the scale of each query direction.
We apply diagonal-pivoted Cholesky to $G$ and retain the selected pivot
positions for residual calibration.
The construction and its equivalence to column-pivoted QR (CPQR) are given in
Appendix~\ref{app:key_routing}.
\subsection{Page Selection and Exact Key Retrieval}
\label{sec:page_routing}
The approximate scores in~\eqref{eq:query_residual_score} rank tokens
without scanning the original Key cache.
For each query head, we assign page $\mathcal P_j$ the score
$
\log\sum_{t\in\mathcal P_j}\exp(\widehat s_{ht})
$
and retain the highest-scoring pages under a fixed budget.
The original post-RoPE Keys are then accessed only for the selected pages,
where their QK scores are recomputed exactly.
The resulting exact scores are used to attend to the corresponding
compressed Values.
Thus, full-dimensional Keys need not be read for unselected pages during
decoding and may reside in a slower memory tier.

\section{Experiment Results}
\begingroup
\setlength{\textfloatsep}{8pt plus 2pt minus 2pt}
\setlength{\intextsep}{8pt plus 2pt minus 2pt}
\setlength{\floatsep}{6pt plus 2pt minus 2pt}
\setlength{\columnsep}{10pt}
\setlength{\abovecaptionskip}{4pt}
\setlength{\belowcaptionskip}{0pt}
\newcommand{\finishExperimentWrap}{%
    \par
    \ifnum\value{WF@wrappedlines}>1
        \vspace{\dimexpr\value{WF@wrappedlines}\baselineskip-\baselineskip\relax}%
    \fi
    \WFclear
}
\paragraph{Experimental Setup.}
We evaluate BasisKV on Llama-3.1, Qwen3, and Qwen3.5.
All Value-compression methods use the same 256 C4 calibration sequences
of 2048 tokens. Key routing uses 32 128K-token windows, equally split
between C4 and synthetic long-range retrieval data, plus 16 held-out C4
validation windows. The default router combines a 16-dimensional
Value-derived Base and a 16-dimensional residual (B16R16), with page size 4.
We evaluate routing on RULER-128K~\citep{hsieh2024ruler}
and nine LongBench~\citep{bai2024longbench} tasks with inputs longer than 4K tokens, selecting 2048 and 256 routed tokens, respectively,
plus 64 recent tokens. Appendix~\ref{app:exp_setup} details calibration,
baselines, and measurement protocols.

\noindent\begin{minipage}{\linewidth}
\begin{wrapfigure}{r}{0.40\textwidth}
    \centering
    \setlength{\abovecaptionskip}{3pt}
    \setlength{\belowcaptionskip}{0pt}
    \includegraphics[width=\linewidth]{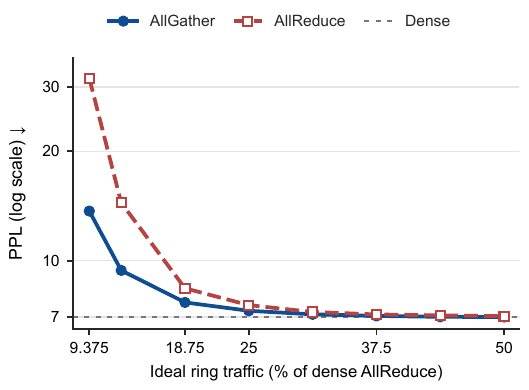}
    \caption{\textbf{Perplexity at matched TP traffic.}
    Qwen3-8B-Base, WikiText-2, TP4.
    Gray dotted line: dense quality.}
    \label{fig:collective-choice-ppl-main}
\end{wrapfigure}
\paragraph{Collective Choice.}
Figure~\ref{fig:collective-choice-ppl-main} compares AllGather and
AllReduce on Qwen3-8B-Base under 4-way tensor parallelism (TP=4)
at matched ideal ring traffic.
Both are jointly fitted against the full attention output. AllGather retains
source-specific latents and decoders; AllReduce sums latents before a shared decoder.

AllGather achieves lower WikiText-2 perplexity throughout, especially at
low traffic. Both approach dense quality as traffic increases, supporting
source-specific latents under constrained communication.
\paragraph{Value Compression across Models.}
Table~\ref{tab:v-quality} compares BasisKV with PaLU M-LRD and G-LRD
at matched global Value-retention budgets of 75\% and 50\%.
Each method uses its own rank-allocation procedure.
M-LRD is the matched-efficiency PaLU baseline; G-LRD groups 4 heads
for stronger quality at $4\times$ the low-rank decoding cost.
BasisKV achieves the highest average accuracy among compressed methods
on all three models at both budgets, while reducing $W_O$ communication.
It also gives the lowest WikiText-2 and C4 perplexity on both Qwen3 models;
PaLU G-LRD is more competitive on Llama-3.1-8B-Instruct.
\finishExperimentWrap
\end{minipage}

\begin{table}[!h]
\tiny
\centering
\setlength{\abovecaptionskip}{0pt}
\setlength{\belowcaptionskip}{3pt}
\caption{\textbf{Value compression across models.}
PaLU M-LRD uses per-head decomposition; G-LRD groups 4 heads.
Comm. is $W_O$ traffic relative to dense.
Bold marks the best compressed result within each budget.}
\label{tab:v-quality}

\renewcommand{\arraystretch}{1.0} 
\setlength{\tabcolsep}{1.1pt}     
\setlength{\aboverulesep}{0.3ex}
\setlength{\belowrulesep}{0.3ex}
\resizebox{\textwidth}{!}{%
\begin{tabular}{@{}lll!{\vrule width 0.45pt}cc!{\vrule width 0.45pt}ccccccc!{\vrule width 0.45pt}c!{\vrule width 0.45pt}c@{}}
\toprule
& & & \multicolumn{2}{c!{\vrule width 0.45pt}}{\textbf{PPL} $\downarrow$} & \multicolumn{7}{c!{\vrule width 0.45pt}}{\textbf{General MCQ} $\uparrow$} & \textbf{MCQ} $\uparrow$ & \textbf{Comm.} $\downarrow$ \\
\cmidrule(lr){4-5}\cmidrule(lr){6-12}\cmidrule(lr){13-13}\cmidrule(l){14-14}
Model & Budget & Method & WikiT. & C4 & BoolQ & ARC-E & ARC-C & WinoG. & PIQA & HellaS. & OB & Avg. & Relative \\
\midrule
\rowcolor{gray!10}
\multirow{7}{*}{\shortstack[l]{Qwen3-8B\\Base}} & \multicolumn{2}{l!{\vrule width 0.45pt}}{\textbf{Dense}} & 7.00 & 9.17 & 0.831 & 0.802 & 0.569 & 0.728 & 0.794 & 0.786 & 0.420 & 0.704 & 1.00 \\
\cmidrule(lr){2-14}
 & \multirow{3}{*}{75\%} & PaLU M-LRD & 8.66 & 10.75 & 0.794 & 0.740 & 0.522 & 0.719 & 0.779 & 0.768 & 0.394 & 0.674 & 1.00 \\
 &  & PaLU G-LRD & 8.24 & 10.40 & 0.793 & 0.771 & \textbf{0.556} & 0.718 & \textbf{0.796} & \textbf{0.787} & 0.408 & 0.690 & 1.00 \\
 &  & BasisKV & \textbf{7.28} & \textbf{9.34} & \textbf{0.818} & \textbf{0.788} & 0.544 & \textbf{0.730} & 0.793 & 0.772 & \textbf{0.424} & \textbf{0.696} & \textbf{0.375} \\
\cmidrule(lr){2-14}
 & \multirow{3}{*}{50\%} & PaLU M-LRD & 14.84 & 16.87 & 0.763 & 0.626 & 0.421 & 0.670 & 0.739 & 0.700 & 0.356 & 0.611 & 1.00 \\
 &  & PaLU G-LRD & 10.65 & 12.87 & 0.720 & 0.698 & 0.502 & 0.710 & 0.767 & \textbf{0.754} & \textbf{0.406} & 0.651 & 1.00 \\
 &  & BasisKV & \textbf{8.26} & \textbf{10.06} & \textbf{0.804} & \textbf{0.771} & \textbf{0.519} & \textbf{0.719} & \textbf{0.788} & 0.735 & 0.398 & \textbf{0.676} & \textbf{0.250} \\
\midrule
\rowcolor{gray!10}
\multirow{7}{*}{\shortstack[l]{Llama-3.1\\8B-Instruct}} & \multicolumn{2}{l!{\vrule width 0.45pt}}{\textbf{Dense}} & 7.22 & 9.77 & 0.841 & 0.796 & 0.552 & 0.741 & 0.810 & 0.792 & 0.430 & 0.709 & 1.00 \\
\cmidrule(lr){2-14}
 & \multirow{3}{*}{75\%} & PaLU M-LRD & 8.60 & 11.25 & 0.832 & 0.778 & 0.528 & 0.747 & \textbf{0.795} & 0.775 & 0.430 & 0.698 & 1.00 \\
 &  & PaLU G-LRD & \textbf{7.98} & \textbf{10.66} & 0.837 & 0.779 & 0.529 & 0.736 & 0.792 & \textbf{0.782} & 0.430 & 0.698 & 1.00 \\
 &  & BasisKV & 8.61 & 10.66 & \textbf{0.846} & \textbf{0.787} & \textbf{0.534} & \textbf{0.748} & \textbf{0.795} & 0.776 & \textbf{0.440} & \textbf{0.704} & \textbf{0.375} \\
\cmidrule(lr){2-14}
 & \multirow{3}{*}{50\%} & PaLU M-LRD & 16.09 & 17.16 & 0.725 & 0.597 & 0.364 & 0.702 & 0.738 & 0.679 & 0.376 & 0.597 & 1.00 \\
 &  & PaLU G-LRD & \textbf{10.61} & 12.90 & 0.813 & 0.712 & 0.462 & 0.710 & 0.764 & 0.738 & 0.396 & 0.656 & 1.00 \\
 &  & BasisKV & 11.50 & \textbf{12.45} & \textbf{0.831} & \textbf{0.735} & \textbf{0.490} & \textbf{0.721} & \textbf{0.773} & \textbf{0.739} & \textbf{0.422} & \textbf{0.673} & \textbf{0.250} \\
\midrule
\rowcolor{gray!10}
\multirow{7}{*}{\shortstack[l]{Qwen3\\32B}} & \multicolumn{2}{l!{\vrule width 0.45pt}}{\textbf{Dense}} & 7.61 & 10.81 & 0.865 & 0.830 & 0.606 & 0.739 & 0.822 & 0.826 & 0.460 & 0.735 & 1.00 \\
\cmidrule(lr){2-14}
 & \multirow{3}{*}{75\%} & PaLU M-LRD & 8.48 & 11.46 & 0.859 & 0.814 & 0.611 & 0.715 & 0.808 & 0.816 & 0.446 & 0.724 & 1.00 \\
 &  & PaLU G-LRD & 8.24 & 11.27 & \textbf{0.866} & 0.808 & \textbf{0.624} & \textbf{0.736} & 0.805 & 0.819 & 0.448 & 0.729 & 1.00 \\
 &  & BasisKV & \textbf{7.76} & \textbf{10.92} & 0.864 & \textbf{0.832} & 0.623 & 0.716 & \textbf{0.808} & \textbf{0.825} & \textbf{0.480} & \textbf{0.735} & \textbf{0.600} \\
\cmidrule(lr){2-14}
 & \multirow{3}{*}{50\%} & PaLU M-LRD & 10.29 & 13.35 & 0.742 & 0.740 & 0.521 & 0.710 & 0.785 & 0.769 & 0.430 & 0.671 & 1.00 \\
 &  & PaLU G-LRD & 9.16 & 12.07 & \textbf{0.874} & 0.749 & 0.574 & \textbf{0.740} & 0.793 & 0.795 & 0.434 & 0.708 & 1.00 \\
 &  & BasisKV & \textbf{8.62} & \textbf{11.62} & 0.841 & \textbf{0.809} & \textbf{0.593} & 0.715 & \textbf{0.807} & \textbf{0.808} & \textbf{0.446} & \textbf{0.717} & \textbf{0.400} \\
\bottomrule
\end{tabular}
}
\par
\vspace{4pt}

\end{table}

\FloatBarrier

\paragraph{Tensor-Parallel Inference.}
Communication-compression gains grow with batch size on both models
(Figure~\ref{fig:tp8-serving-speedup}). V64 crosses over earlier and
provides larger gains than V96, reaching $1.42\times$ end-to-end speedup
on Llama-3.1-8B-Instruct and $1.31\times$ on Qwen3-32B at batch size 256.
Time-to-first-token (TTFT) follows the same trend.
\begin{figure}[!ht]
    \centering
    \includegraphics[width=0.94\linewidth]{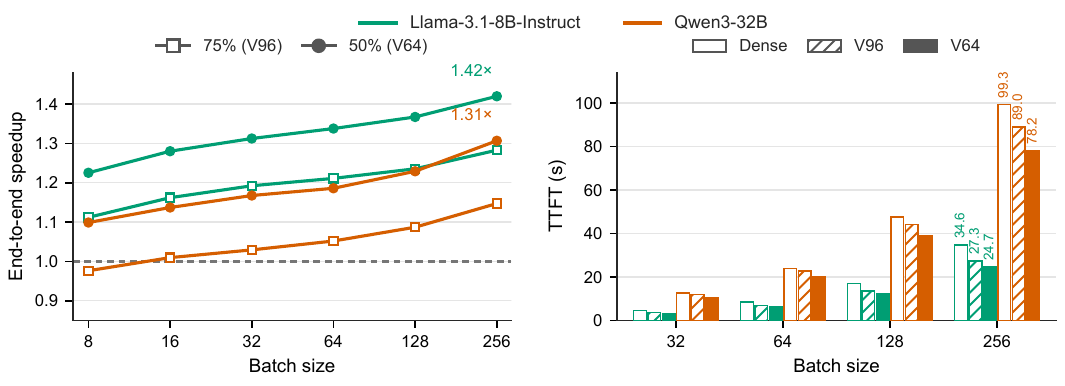}
    \caption{\textbf{TP inference.}
Llama-3.1-8B-Instruct and Qwen3-32B, $8\times$ L40S;
4K-token prompts, 128 output tokens. Left: end-to-end speedup;
right: TTFT. V96/V64 retain Value widths of 96/64.}
    \label{fig:tp8-serving-speedup}
\end{figure}
\FloatBarrier

\paragraph{Long-Context Routing Quality.}
In Table~\ref{tab:ruler-longbench}, Full-K uses full attention. Support is approximately matched for LRQK and Loki as described in Appendix C.1, while ShadowKV retains its native support.
BasisKV remains close to Full-K on RULER and among the strongest
sparse methods. Its retrieval results support combining Value-derived
prediction with residual correction. LongBench differences are smaller
and model-dependent.

\begin{table}[!ht]
\centering
\tiny
\setlength{\abovecaptionskip}{0pt}
\caption{\textbf{Sparse attention across models.}
RULER (10 tasks) and LongBench (9 tasks).
Avg. is the unweighted task mean. Bold marks the best sparse score.}
\label{tab:ruler-longbench}
\renewcommand{\arraystretch}{1.15}
\setlength{\tabcolsep}{1.3pt}
\resizebox{\textwidth}{!}{%
\begin{tabular}{@{}ll!{\vrule width 0.45pt}*{10}{r}!{\vrule width 0.45pt}r!{\vrule width 0.45pt}*{9}{r}!{\vrule width 0.45pt}r@{}}
\toprule
& & \multicolumn{11}{c!{\vrule width 0.45pt}}{\textbf{RULER} $\uparrow$} & \multicolumn{10}{c}{\textbf{LongBench} $\uparrow$} \\
\cmidrule(lr){3-13}\cmidrule(lr){14-23}
Model & Method & S1 & S2 & MK1 & MK2 & MQ & MV & QA1 & QA2 & VT & FWE & Avg. & NQA & MQA & HQA & MQue & DRead & GRep & SAM & PRetr & LCC & Avg. \\
\midrule
\rowcolor{gray!10}
\multirow{5}{*}{\shortstack[l]{Llama-3.1\\8B-Instruct}} & Full-K & 100.00 & 100.00 & 99.00 & 83.00 & 99.00 & 95.00 & 74.00 & 48.00 & 76.80 & 52.67 & 82.75 & 28.06 & 50.25 & 52.23 & 31.32 & 32.34 & 32.76 & 44.50 & 99.50 & 52.22 & 47.02 \\
\cmidrule(lr){2-23}
 & Loki & 79.00 & 96.00 & 97.00 & 18.00 & 55.50 & 53.25 & 48.00 & 32.00 & 25.00 & 48.33 & 55.21 & 27.72 & 48.48 & 52.04 & 31.23 & 29.54 & \textbf{32.36} & \textbf{46.06} & 99.50 & 51.95 & 46.54 \\
 & ShadowKV & 99.00 & 99.00 & 98.00 & 61.00 & 97.00 & 91.75 & 74.00 & \textbf{48.00} & 28.40 & \textbf{58.00} & 75.41 & \textbf{29.22} & \textbf{51.06} & \textbf{52.64} & \textbf{31.49} & \textbf{31.12} & 30.99 & 43.74 & 99.00 & 51.18 & 46.72 \\
 & LRQK & 100.00 & 100.00 & \textbf{99.00} & 76.00 & 99.00 & 93.25 & 75.00 & 47.00 & 53.60 & 48.67 & 79.15 & 27.90 & 50.00 & 52.13 & 31.46 & 30.63 & 31.00 & 45.37 & 99.50 & 51.61 & 46.62 \\
 & BasisKV & \textbf{100.00} & \textbf{100.00} & 98.00 & \textbf{78.00} & \textbf{99.25} & \textbf{94.50} & \textbf{75.00} & 47.00 & \textbf{64.80} & 46.67 & \textbf{80.32} & 29.05 & 51.02 & 52.51 & 30.70 & 28.95 & 31.47 & 44.90 & \textbf{99.50} & \textbf{54.05} & \textbf{46.91} \\
\midrule
\rowcolor{gray!10}
\multirow{5}{*}{\shortstack[l]{Qwen3-8B}} & Full-K & 100.00 & 100.00 & 85.00 & 61.00 & 93.50 & 96.50 & 55.00 & 36.00 & 93.40 & 88.00 & 80.84 & 25.35 & 51.47 & 51.61 & 27.34 & 17.89 & 33.53 & 46.52 & 100.00 & 62.06 & 46.20 \\
\cmidrule(lr){2-23}
 & Loki & 100.00 & 99.00 & 85.00 & 38.00 & 91.00 & 90.75 & 43.00 & 27.00 & 78.40 & 73.33 & 72.55 & 26.47 & \textbf{51.69} & 51.11 & \textbf{27.69} & \textbf{22.58} & 32.94 & 44.94 & 98.00 & 55.09 & 45.61 \\
 & ShadowKV & 100.00 & 95.00 & 84.00 & 44.00 & 83.75 & 86.75 & 53.00 & 28.00 & 82.00 & \textbf{88.67} & 74.52 & 26.12 & 50.47 & 51.94 & 27.32 & 19.93 & \textbf{33.82} & 44.67 & 91.00 & 55.10 & 44.49 \\
 & LRQK$^{\dagger}$ & 100.00 & 98.00 & 86.00 & \textbf{57.00} & 93.75 & \textbf{96.50} & 51.00 & \textbf{36.00} & 91.20 & 81.33 & \textbf{79.08} & 26.76 & 49.50 & \textbf{52.93} & 26.86 & 16.36 & 33.22 & \textbf{46.64} & \textbf{100.00} & \textbf{59.84} & \textbf{45.79} \\
 & BasisKV & \textbf{100.00} & \textbf{99.00} & \textbf{86.00} & 53.00 & \textbf{94.00} & 94.00 & \textbf{54.00} & 33.00 & \textbf{93.00} & 80.33 & 78.63 & \textbf{26.97} & 50.94 & 51.47 & 26.63 & 16.68 & 32.86 & 44.60 & 98.50 & 58.80 & 45.27 \\
\midrule
\rowcolor{gray!10}
\multirow{5}{*}{\shortstack[l]{Qwen3.5-9B}} & Full-K & 100.00 & 100.00 & 100.00 & 100.00 & 100.00 & 100.00 & 86.00 & 56.00 & 100.00 & 100.00 & 94.20 & 33.85 & 53.95 & 61.71 & 47.02 & 24.61 & 31.04 & 24.27 & 29.42 & 65.80 & 41.30 \\
\cmidrule(lr){2-23}
 & Loki & 100.00 & 100.00 & 100.00 & 97.00 & 99.75 & 100.00 & 80.00 & 55.00 & 100.00 & 98.67 & 93.04 & 32.52 & 54.51 & 61.27 & 47.14 & 22.96 & \textbf{31.10} & 24.45 & 30.17 & 64.15 & 40.92 \\
 & ShadowKV & 100.00 & 99.00 & 99.00 & 100.00 & 99.00 & 95.00 & 84.00 & \textbf{58.00} & 94.80 & \textbf{99.67} & 92.85 & \textbf{33.14} & \textbf{54.96} & \textbf{61.87} & 46.46 & 23.09 & 31.04 & 23.84 & \textbf{40.75} & 61.35 & \textbf{41.83} \\
 & LRQK & 100.00 & 100.00 & 100.00 & 100.00 & 100.00 & 98.75 & \textbf{85.00} & 55.00 & 99.20 & 99.00 & 93.70 & 32.89 & 53.20 & 61.85 & \textbf{47.15} & 23.98 & 30.90 & \textbf{24.64} & 30.92 & \textbf{64.17} & 41.08 \\
 & BasisKV & \textbf{100.00} & \textbf{100.00} & \textbf{100.00} & \textbf{100.00} & \textbf{100.00} & \textbf{100.00} & 83.00 & 56.00 & \textbf{100.00} & 98.00 & \textbf{93.70} & 33.02 & 54.29 & 61.00 & 46.31 & \textbf{24.78} & 30.74 & 23.77 & 35.08 & 62.87 & 41.32 \\
\bottomrule
\end{tabular}%
}

\end{table}

\FloatBarrier

\begin{table}[!ht]
    \centering
    \begin{minipage}{0.70\textwidth}
    \centering
    \setlength{\abovecaptionskip}{0pt}
    \setlength{\belowcaptionskip}{4pt}
    \caption{\textbf{Sparse decoding across context lengths.}
    TP1/B1, one L40S; median ms over three trials.
    $\times$: speedup over dense.}
    \label{tab:tp1-gpu-local-scaling}
    \footnotesize
    \setlength{\tabcolsep}{2.5pt}
    \renewcommand{\arraystretch}{1.0}
    \begin{tabular*}{\linewidth}{@{\extracolsep{\fill}}crrrrrr@{}}
        \toprule
        & \multicolumn{3}{c}{Attention block}
        & \multicolumn{3}{c}{Full-model decoding} \\
        \cmidrule(lr){2-4}\cmidrule(l){5-7}
        Context & Dense & BasisKV & $\times$ & Dense & BasisKV & $\times$ \\
        \midrule
        24K & 5.27 & 5.17 & 1.02 & 28.63 & 28.33 & 1.01 \\
        32K & 6.70 & 5.85 & 1.15 & 30.11 & 29.00 & 1.04 \\
        48K & 9.61 & 7.31 & 1.31 & 33.02 & 30.43 & 1.09 \\
        64K & 12.58 & 8.70 & 1.45 & 36.01 & 31.93 & 1.13 \\
        96K & 18.33 & 11.77 & 1.56 & 41.78 & 34.98 & 1.19 \\
        128K & 23.85 & 14.67 & 1.63 & 47.31 & 37.87 & 1.25 \\
        \bottomrule
    \end{tabular*}
    \end{minipage}
\end{table}
\paragraph{Sparse Decoding.}
Table~\ref{tab:tp1-gpu-local-scaling} evaluates B16R16 at 24K--128K,
excluding prefill. Steady-state attention-block and full-model speedups
grow with context, reaching $1.63\times$ and $1.25\times$ at 128K.
All exact Keys and Values remain on GPU, without CPU offload.

\paragraph{Long-Context Throughput and Memory.}
Figure~\ref{fig:tp8-throughput-memory} shows throughput gains across batch
sizes and context lengths, reaching $2.06\times$ for Llama-3.1-8B-Instruct
and $1.99\times$ for Qwen3-8B at comparable measured decode memory footprints.

\begin{figure}[!t]
    \centering
    \includegraphics[width=0.94\linewidth]{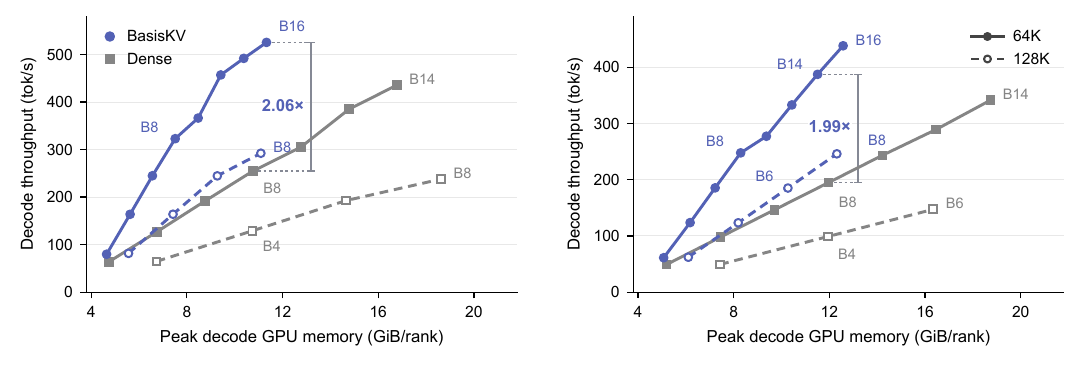}
    \caption{\textbf{Decode throughput and GPU memory.}
    Llama-3.1-8B-Instruct (left) and Qwen3-8B (right), TP8 on $8\times$ L40S.
    Memory: peak per-rank PyTorch allocation during decode.
    Circles/squares: BasisKV/dense; solid/dashed: 64K/128K.}
    \label{fig:tp8-throughput-memory}
\end{figure}
\FloatBarrier
\endgroup

\section{Related Work}
\paragraph{Tensor-parallel communication.}
Tensor parallelism partitions Transformer operators across accelerators and
uses collective communication to combine their partial
results~\citep{shoeybi2019megatron}.
Prior work reduces this communication through activation
quantization~\citep{hansenpalmus2024communication} or improves parallel
execution for decomposed models~\citep{huang2026deinfer}.
BasisKV instead preserves the individual TP sources through a low-rank
AllGather and jointly fits their factors against the complete attention
output. The resulting Value representation is also reused for caching and
Key routing.

\paragraph{KV-cache compression.}
KV-cache compression has been studied through quantization
\citep{liu2024kivi,hooper2024kvquant,kang2024gear},
token and head eviction
\citep{zhang2023h2o,ge2024fastgen,li2024snapkv,
cai2024pyramidkv,xiao2025duoattention,tang2025razorattention},
and low-dimensional representations
\citep{chang2025palu,lin2025matryoshkakv,bhatnagar2026starkv}.
PaLU stores low-rank latent Key and Value states obtained from
decomposed projection matrices and absorbs Value reconstruction into
the output projection~\citep{chang2025palu}.
STAR-KV learns adaptive low-rank truncation through differentiable
soft-thresholding and additional distillation-based training
\citep{bhatnagar2026starkv}.
BasisKV instead derives its Value representation post hoc by fitting
the complete TP attention output, so the cached representation is
simultaneously optimized for the communication path.

\paragraph{Sparse attention and KV routing.}
Sparse-attention methods reduce decode-time memory traffic by selecting a
subset of cached states for attention
\citep{zhang2023h2o,li2024snapkv,cai2024pyramidkv}.
SparQ selectively fetches cached history~\citep{ribar2024sparq}, while
Quest scores KV-cache pages using query-dependent Key
summaries~\citep{tang2024quest}.
Loki ranks tokens using a low-dimensional Key
representation~\citep{singhania2024loki}, and ShadowKV retains a low-rank
Key representation on GPU while retrieving selected KV
states~\citep{sun2025shadowkv}.
Recent low-rank approaches similarly construct compact Query--Key
representations for sparse retrieval, including
LRQK~\citep{li2025lrqk} and SALS~\citep{mu2025sals}.
BasisKV instead reuses routing information already present in the compressed
Value representation and combines it with a Key-specific residual.

\section{Conclusion}

We presented BasisKV, a low-rank KV-cache compression method for
communication-intensive long-context LLM inference. BasisKV jointly
fits low-rank factors against the complete tensor-parallel attention
output, allowing the resulting Value representation to reduce both
TP communication and Value-cache width. The same Value latent is
further reused for Key routing, with a Key-specific residual
capturing information not recovered from Value. Across multiple model
families and system settings, BasisKV improves the quality--efficiency
tradeoff of low-rank compression while providing end-to-end gains in
both tensor-parallel and long-context sparse decoding. These results
show that a shared Value representation can serve multiple inference
boundaries without optimizing each component in isolation.

\bibliography{reference}

@article{shoeybi2019megatron,
  title={Megatron-LM: Training Multi-Billion Parameter Language Models Using Model Parallelism},
  author={Shoeybi, Mohammad and Patwary, Mostofa and Puri, Raul and LeGresley, Patrick and Casper, Jared and Catanzaro, Bryan},
  journal={arXiv preprint arXiv:1909.08053},
  year={2019}
}

@article{hansenpalmus2024communication,
  title={Communication Compression for Tensor Parallel LLM Inference},
  author={Hansen-Palmus, Jan and Le, Michael Truong and Hausd{\"o}rfer, Oliver and Verma, Alok},
  journal={arXiv preprint arXiv:2411.09510},
  year={2024}
}

@article{huang2026deinfer,
  title={DeInfer: Efficient Parallel Inferencing for Decomposed Large Language Models},
  author={Huang, You-Liang and Huang, Xinhao and Liao, Chengxi and Wen, Zeyi},
  journal={arXiv preprint arXiv:2604.17709},
  year={2026}
}

@inproceedings{liu2024kivi,
  title={{KIVI}: A Tuning-Free Asymmetric 2bit Quantization for {KV} Cache},
  author={Liu, Zirui and Yuan, Jiayi and Jin, Hongye and Zhong, Shaochen and Xu, Zhaozhuo and Braverman, Vladimir and Chen, Beidi and Hu, Xia},
  booktitle={Proceedings of the 41st International Conference on Machine Learning},
  pages={32332--32344},
  year={2024},
  volume={235},
  series={Proceedings of Machine Learning Research},
  publisher={PMLR}
}

@inproceedings{chang2025palu,
  title={Palu: {KV}-Cache Compression with Low-Rank Projection},
  author={Chang, Chi-Chih and Lin, Wei-Cheng and Lin, Chien-Yu and Chen, Chong-Yan and Hu, Yu-Fang and Wang, Pei-Shuo and Huang, Ning-Chi and Ceze, Luis and Abdelfattah, Mohamed S. and Wu, Kai-Chiang},
  booktitle={International Conference on Learning Representations},
  year={2025}
}

@inproceedings{ribar2024sparq,
  title={{SparQ} Attention: Bandwidth-Efficient {LLM} Inference},
  author={Ribar, Luka and Chelombiev, Ivan and Hudlass-Galley, Luke and Blake, Charlie and Luschi, Carlo and Orr, Douglas},
  booktitle={Proceedings of the 41st International Conference on Machine Learning},
  pages={42558--42583},
  year={2024},
  volume={235},
  series={Proceedings of Machine Learning Research},
  publisher={PMLR}
}

@inproceedings{tang2024quest,
  title={{QUEST}: Query-Aware Sparsity for Efficient Long-Context {LLM} Inference},
  author={Tang, Jiaming and Zhao, Yilong and Zhu, Kan and Xiao, Guangxuan and Kasikci, Baris and Han, Song},
  booktitle={Proceedings of the 41st International Conference on Machine Learning},
  pages={47901--47911},
  year={2024},
  volume={235},
  series={Proceedings of Machine Learning Research},
  publisher={PMLR}
}

@inproceedings{singhania2024loki,
  title={Loki: Low-rank Keys for Efficient Sparse Attention},
  author={Singhania, Prajwal and Singh, Siddharth and He, Shwai and Feizi, Soheil and Bhatele, Abhinav},
  booktitle={Advances in Neural Information Processing Systems},
  pages={16692--16723},
  volume={37},
  year={2024}
}

@inproceedings{sun2025shadowkv,
  title={{ShadowKV}: {KV} Cache in Shadows for High-Throughput Long-Context {LLM} Inference},
  author={Sun, Hanshi and Chang, Li-Wen and Bao, Wenlei and Zheng, Size and Zheng, Ningxin and Liu, Xin and Dong, Harry and Chi, Yuejie and Chen, Beidi},
  booktitle={Proceedings of the 42nd International Conference on Machine Learning},
  pages={57355--57373},
  year={2025},
  volume={267},
  series={Proceedings of Machine Learning Research},
  publisher={PMLR}
}

@inproceedings{li2025lrqk,
  title={Efficient Low Rank Attention for Long-Context Inference in Large Language Models},
  author={Li, Tenghui and Zhou, Guoxu and Zhao, Xuyang and Qiu, Yuning and Zhao, Qibin},
  booktitle={Advances in Neural Information Processing Systems},
  volume={38},
  pages={43048--43074},
  year={2025}
}

@inproceedings{mu2025sals,
  title={{SALS}: Sparse Attention in Latent Space for {KV} Cache Compression},
  author={Mu, Junlin and Huang, Hantao and Zhang, Jihang and Yu, Minghui and Wang, Tao and Li, Yidong},
  booktitle={Advances in Neural Information Processing Systems},
  volume={38},
  pages={428--447},
  year={2025}
}

@article{hsieh2024ruler,
  title={{RULER}: What's the Real Context Size of Your Long-Context Language Models?},
  author={Hsieh, Cheng-Ping and Sun, Simeng and Kriman, Samuel and Acharya, Shantanu and Rekesh, Dima and Jia, Fei and Zhang, Yang and Ginsburg, Boris},
  journal={arXiv preprint arXiv:2404.06654},
  year={2024}
}

@inproceedings{bai2024longbench,
  title={LongBench: A Bilingual, Multitask Benchmark for Long Context Understanding},
  author={Bai, Yushi and Lv, Xin and Zhang, Jiajie and Lyu, Hongchang and Tang, Jiankai and Huang, Zhidian and Du, Zhengxiao and Liu, Xiao and Zeng, Aohan and Hou, Lei and Dong, Yuxiao and Tang, Jie and Li, Juanzi},
  booktitle={Proceedings of the 62nd Annual Meeting of the Association for Computational Linguistics (Volume 1: Long Papers)},
  pages={3119--3137},
  year={2024},
  publisher={Association for Computational Linguistics}
}

@inproceedings{hooper2024kvquant,
  title={{KVQuant}: Towards 10 Million Context Length {LLM} Inference with {KV} Cache Quantization},
  author={Hooper, Coleman and Kim, Sehoon and Mohammadzadeh, Hiva and Mahoney, Michael W. and Shao, Yakun Sophia and Keutzer, Kurt and Gholami, Amir},
  booktitle={Advances in Neural Information Processing Systems},
  volume={37},
  year={2024}
}

@article{kang2024gear,
  title={{GEAR}: An Efficient {KV} Cache Compression Recipe for Near-Lossless Generative Inference of {LLM}},
  author={Kang, Hao and Zhang, Qingru and Kundu, Souvik and Jeong, Geonhwa and Liu, Zaoxing and Krishna, Tushar and Zhao, Tuo},
  journal={arXiv preprint arXiv:2403.05527},
  year={2024}
}

@inproceedings{zhang2023h2o,
  title={{H2O}: Heavy-Hitter Oracle for Efficient Generative Inference of Large Language Models},
  author={Zhang, Zhenyu and Sheng, Ying and Zhou, Tianyi and Chen, Tianlong and Zheng, Lianmin and Cai, Ruisi and Song, Zhao and Tian, Yuandong and R{\'e}, Christopher and Barrett, Clark and Wang, Zhangyang and Chen, Beidi},
  booktitle={Advances in Neural Information Processing Systems},
  volume={36},
  pages={34661--34710},
  year={2023}
}

@inproceedings{ge2024fastgen,
  title={Model Tells You What to Discard: Adaptive {KV} Cache Compression for {LLM}s},
  author={Ge, Suyu and Zhang, Yunan and Liu, Liyuan and Zhang, Minjia and Han, Jiawei and Gao, Jianfeng},
  booktitle={International Conference on Learning Representations},
  year={2024}
}

@inproceedings{li2024snapkv,
  title={{SnapKV}: {LLM} Knows What You Are Looking for Before Generation},
  author={Li, Yuhong and Huang, Yingbing and Yang, Bowen and Venkitesh, Bharat and Locatelli, Acyr and Ye, Hanchen and Cai, Tianle and Lewis, Patrick and Chen, Deming},
  booktitle={Advances in Neural Information Processing Systems},
  volume={37},
  year={2024}
}

@article{cai2024pyramidkv,
  title={{PyramidKV}: Dynamic {KV} Cache Compression based on Pyramidal Information Funneling},
  author={Cai, Zefan and Zhang, Yichi and Gao, Bofei and Liu, Yuliang and Liu, Tianyu and Lu, Keming and Xiong, Wayne and Dong, Yue and Chang, Baobao and Hu, Junjie and Xiao, Wen},
  journal={arXiv preprint arXiv:2406.02069},
  year={2024}
}

@inproceedings{xiao2025duoattention,
  title={{DuoAttention}: Efficient Long-Context {LLM} Inference with Retrieval and Streaming Heads},
  author={Xiao, Guangxuan and Tang, Jiaming and Zuo, Jingwei and Guo, Junxian and Yang, Shang and Tang, Haotian and Fu, Yao and Han, Song},
  booktitle={International Conference on Learning Representations},
  year={2025}
}

@inproceedings{tang2025razorattention,
  title={{RazorAttention}: Efficient {KV} Cache Compression Through Retrieval Heads},
  author={Tang, Hanlin and Lin, Yang and Lin, Jing and Han, Qingsen and Ke, Danning and Hong, Shikuan and Yao, Yiwu and Wang, Gongyi},
  booktitle={International Conference on Learning Representations},
  year={2025}
}

@inproceedings{lin2025matryoshkakv,
  title={{MatryoshkaKV}: Adaptive {KV} Compression via Trainable Orthogonal Projection},
  author={Lin, Bokai and Zeng, Zihao and Xiao, Zipeng and Kou, Siqi and Hou, TianQi and Gao, Xiaofeng and Zhang, Hao and Deng, Zhijie},
  booktitle={International Conference on Learning Representations},
  year={2025}
}

@inproceedings{saxena2024eigen,
  title={Eigen attention: Attention in low-rank space for kv cache compression},
  author={Saxena, Utkarsh and Saha, Gobinda and Choudhary, Sakshi and Roy, Kaushik},
  booktitle={Findings of the Association for Computational Linguistics: EMNLP 2024},
  pages={15332--15344},
  year={2024}
}

@inproceedings{chang2026xkv,
  title={x{KV}: Cross-Layer {KV}-Cache Compression via Aligned Singular Vector Extraction},
  author={Chi-Chih Chang and Wei-Cheng Lin and Chien-Yu Lin and Hung-Yueh Chiang and Yash Akhauri and Xilai Dai and Huiqiang Jiang and Yucheng Li and Luis Ceze and Kai-Chiang Wu and Mohamed S. Abdelfattah},
  booktitle={Forty-third International Conference on Machine Learning},
  year={2026}
}

@article{bhatnagar2026starkv,
  title={{STAR-KV}: Low-Rank {KV} Cache Compression via Soft Thresholding
         for Adaptive Rank Control},
  author={Bhatnagar, Priyansh and Moradifirouzabadi, Ashkan and
          Yang, Se-Hyun and Lee, SeungJae and Choi, Jungwook and
          Kang, Mingu},
  journal={arXiv preprint arXiv:2606.08382},
  year={2026}
}
\bibliographystyle{iclr2027_conference}

\appendix
\section{Appendix}
\label{app:value_compression}
\subsection{Collective-Aware Fitting}
\label{app:als}
For the calibration set, let $Z_h$ denote the activation matrix of source
block $h$, and let $Y$ denote the corresponding complete attention output.
For compactness, define the Gram and cross-product statistics
\begin{equation*}
    C_{hk}=Z_h^\top Z_k,
    \qquad
    B_h=Z_h^\top Y.
\end{equation*}
\paragraph{Decoder update.}
With the encoders fixed, the decoder blocks are solved jointly.
For each $h$, stationarity with respect to $D_h$ gives
\begin{align*}
    0
    &=
    E_h^\top Z_h^\top
    \left(
        \sum_{k=1}^{H}
        Z_kE_kD_k
        -
        Y
    \right) =
    \sum_{k=1}^{H}
    E_h^\top
    C_{hk}
    E_k
    D_k
    -
    E_h^\top B_h.
\end{align*}
Thus, the $h$ th stationarity condition couples $D_h$ to all decoder
blocks $\{D_k\}_{k=1}^H$.
The resulting normal equations are
\begin{equation}
    \sum_{k=1}^{H}
    \left(
        E_h^\top C_{hk}E_k
    \right)
    D_k
    =
    E_h^\top B_h,
    \qquad
    h=1,\ldots,H.
    \label{eq:app_decoder_normal}
\end{equation}
We assemble this coupled block system from the calibration statistics and
solve for all decoder blocks jointly by Cholesky factorization.

\paragraph{Encoder update.}
With the decoders fixed, we update one encoder at a time.
For encoder $E_h$, we first subtract the contributions of all other
source blocks and define the conditional target
\begin{equation*}
    T_h
    =
    Y
    -
    \sum_{k\ne h}
    Z_kE_kD_k.
\end{equation*}
With the remaining encoders fixed, the joint objective therefore reduces
to the least-squares problem
\begin{equation*}
    \min_{E_h}
    \left\|
        T_h-Z_hE_hD_h
    \right\|_F^2.
\end{equation*}
Its first-order optimality condition gives the normal equation
\begin{equation}
    C_{hh}E_hD_hD_h^\top
    =
    Z_h^\top T_hD_h^\top.
    \label{eq:app_encoder_normal}
\end{equation}
Using the definition of $T_h$, the right-hand side can be written entirely
in terms of the calibration statistics as
\begin{equation*}
    C_{hh}E_hD_hD_h^\top
    =
    B_hD_h^\top
    -
    \sum_{k\ne h}
    C_{hk}E_kD_kD_h^\top.
\end{equation*}
The left-hand side defines the linear operator
$
    \mathcal H_h(X)
    =
    C_{hh}X D_hD_h^\top.
$
It is positive semidefinite since
\begin{equation*}
    \langle X,\mathcal H_h(X)\rangle_F
    =
    \|Z_hXD_h\|_F^2
    \ge 0.
\end{equation*}
Vectorizing \eqref{eq:app_encoder_normal} would produce a
Kronecker-structured linear system with coefficient matrix
$
(D_hD_h^\top)\otimes C_{hh}
$.
Rather than form this matrix explicitly, we solve the normal equation by
matrix-free conjugate gradient using
$
\mathcal H_h(X)=C_{hh}XD_hD_h^\top
$
for each matrix--vector product.

Because the factorization is invariant to a change of latent basis, we
orthonormalize each updated encoder by a thin QR factorization
$
E_h=Q_hR_h
$
and set
$
E_h\leftarrow Q_h
$
and
$
D_h\leftarrow R_hD_h.
$
This leaves $E_hD_h$ unchanged and avoids scale drift between the two
factors.
After the final encoder sweep, we solve
\eqref{eq:app_decoder_normal} once more for the decoder blocks.
\begin{algorithm}[!ht]
\caption{Collective-Aware ALS}
\label{alg:collective_als}
\begin{algorithmic}[1]
\Function{Collective-Aware-ALS}{$\{C_{hk}\},\{B_h\},\{E_h\},S$}
    \State Solve~\eqref{eq:app_decoder_normal} for $\{D_h\}$
        \Comment{joint Cholesky solve}
    \For{$s=1,\ldots,S$}
        \For{$h=1,\ldots,H$}
            \State Solve~\eqref{eq:app_encoder_normal} for $E_h$
                \Comment{matrix-free CG}
            \State $[Q_h,R_h]\gets\Call{QR}{E_h}$
            \State $E_h\gets Q_h$
            \State $D_h\gets R_hD_h$
        \EndFor
        \State Solve~\eqref{eq:app_decoder_normal} for $\{D_h\}$
            \Comment{joint Cholesky solve}
    \EndFor
    \State \Return $\{E_h\},\{D_h\}$
\EndFunction
\end{algorithmic}
\end{algorithm}

\subsection{Rank Allocation}
\label{app:rank_allocation}
Let $\ell=1,\ldots,L$ index the compressed layers, and let
$e_\ell(r)$ denote the reconstruction error of layer $\ell$ at rank $r$.
The two-sided perturbations in \eqref{eq:two_sided_kl} relate changes in
$e_\ell(r)$ to changes in end-to-end KL divergence.
Using the measured sensitivities $s_\ell^-$ and $s_\ell^+$, we estimate
the KL change at a candidate rank $r$ by
\begin{equation*}
    \widehat{\Delta\mathcal K}_\ell(r)
    =
    \begin{cases}
        s_\ell^-
        \left[
            e_\ell(r)-e_\ell(r_0)
        \right],
        & r<r_0,
        \\[3pt]
        0,
        & r=r_0,
        \\[3pt]
        s_\ell^+
        \left[
            e_\ell(r)-e_\ell(r_0)
        \right],
        & r>r_0.
    \end{cases}
\end{equation*}
Thus the reconstruction curve of each layer is converted into an
estimated end-to-end KL change using only the two measured perturbations.
We use the separable approximation
\begin{equation*}
    \mathcal K(\mathbf r)-\mathcal K(\mathbf r_0)
    \approx
    \sum_{\ell=1}^{L}
    \widehat{\Delta\mathcal K}_\ell(r_\ell).
\end{equation*}

For a total rank budget $B$, the allocation is therefore
\begin{equation*}
    \min_{r_1,\ldots,r_L\in\mathcal R}
    \sum_{\ell=1}^{L}
    \widehat{\Delta\mathcal K}_\ell(r_\ell)
    \qquad
    \text{subject to}
    \qquad
    \sum_{\ell=1}^{L}r_\ell=B.
\end{equation*}
Let $F(\ell,b)$ denote the minimum estimated KL change for the first
$\ell$ layers with total rank $b$.
We solve the exact-budget allocation by dynamic programming with
\begin{equation*}
    F(\ell,b)
    =
    \min_{\substack{r\in\mathcal R\\r\le b}}
    \left\{
        F(\ell-1,b-r)
        +
        \widehat{\Delta\mathcal K}_\ell(r)
    \right\}.
\end{equation*}
The resulting dynamic program gives the rank schedule under the budget $B$.

\subsection{Architecture Extensions}
\label{app:gated_attention}
\label{app:architecture_extensions}

The output-side BasisKV factorization only requires a linear output
projection.
It therefore applies without modification to other sequence-model blocks
with the same structure, including Gated DeltaNet and Mamba blocks.
In these blocks, we apply BasisKV to the output projection $W_O$ and no
Value-cache factorization is introduced.
\paragraph{Gated attention.}
For standard attention, the Value reconstruction can be reassociated with
the output projection, allowing the jointly fitted factors to also define
the compressed Value cache.
An elementwise output gate breaks this reassociation.
For a gated attention block, we therefore use a separate Value-side
factorization
\begin{equation*}
    \widehat Y
    =
    \sum_{h=1}^{H_q}
    \left[
        \left(
            Z_hE_h^V R_h^V
        \right)
        \odot G_h
    \right]
    W_{O,h},
\end{equation*}
where $G_h$ is the output gate.
The Value-side factors $E_h^V$ and $R_h^V$ are fitted before the gate and
are not part of the joint output-side fit.
The encoder $E_h^V$ is folded into the Value projection, so
$
\widehat V_h=V_hE_h^V
$
is stored in the cache.

After the gate, the resulting activations are treated as the sources of
the output projection, and the standard collective-aware BasisKV
factorization is applied to $W_O$.
Thus gated attention uses a separate Value-cache factorization followed
by the same jointly fitted output-side compression used elsewhere.

\section{Value-Based Key Routing}
\label{app:key_routing}
\subsection{Reduced-Rank Key Prediction}
\label{app:key_prediction}

Let
$
G=\widehat V_c^\top\widehat V_c
$
and
$
H=\widehat V_c^\top K_c
$.
Since
$
\operatorname{range}(H)\subseteq\operatorname{range}(G),
$
the objective in \eqref{eq:value_key_rrr} can be written as
\begin{equation*}
    \|K_c-\widehat V_cM\|_F^2
    =
    \|K_c\|_F^2
    -
    \|G^{\dagger/2}H\|_F^2
    +
    \left\|
        G^{1/2}M-G^{\dagger/2}H
    \right\|_F^2.
\end{equation*}
The first two terms are independent of $M$, so it remains to minimize
\begin{equation*}
    \left\|
        G^{1/2}M-G^{\dagger/2}H
    \right\|_F^2
\end{equation*}
subject to $\operatorname{rank}(M)\le r_b$.
Let
$
U\Sigma V^\top
$
be the rank-$r_b$ truncated singular value decomposition of
$
G^{\dagger/2}H
$.
Since
$
G^{\dagger/2}H
$
lies in the range of $G$, its truncated SVD is attainable as
$
G^{1/2}M
$
and gives the best rank-$r_b$ approximation.
Thus one minimizer is
\begin{equation*}
    M^\star
    =
    G^{\dagger/2}U\Sigma V^\top.
\end{equation*}

For the orthogonal matrix $Q=[Q_b\;Q_\perp]$ defined in
Section~\ref{sec:key_prediction}, we have
$
QQ^\top=I.
$
Hence
\begin{equation*}
    (\widehat VQ)(Q^\top D)
    =
    \widehat VQQ^\top D
    =
    \widehat VD.
\end{equation*}
The change of basis leaves the Value reconstruction unchanged.
Moreover,
\begin{equation*}
    \widehat V_cM^\star
    =
    (\widehat V_cQ)_{:,:r_b}R_bV^\top,
\end{equation*}
so the Key predictor depends only on the first $r_b$ coordinates of the
rotated Value latent.

\subsection{Low-Rank Residual Correction}
\label{app:query_residual}

\paragraph{Residual approximation.}
Let
$
e_t=k_t-\widehat k_t
$
denote the post-RoPE Key residual.
We approximate the exact residual score
$
q_he_t^\top/\sqrt{d_h}
$
by the rank-$r_q$ form
$
(q_hU_h)(e_tF_h)^\top/\sqrt{d_h},
$
where $U_h,F_h\in\mathbb R^{d_h\times r_q}$.
The resulting token-score perturbation is
\begin{equation*}
    \Delta s_{ht}
    =
    \frac{1}{\sqrt{d_h}}
    q_h
    (U_hF_h^\top-I)
    e_t^\top.
\end{equation*}
The corresponding residual representation
$
e_tF_h\in\mathbb R^{r_q}
$
is stored for each cached token.

\subsubsection{Page-Fisher Objective}
\label{app:page_fisher}

Fix one calibration query and head, and suppress these indices.
Let $s_t$ denote the exact token score and
$
p_t=\operatorname{softmax}(s)_t
$ be the corresponding attention probability over the routable prefix.
Since routing is performed over pages, for each page $\mathcal P_p$ we
define its exact attention mass as
\begin{equation*}
    \pi_p
    =
    \sum_{t\in\mathcal P_p} p_t.
\end{equation*}

Let $\Delta s_t$ denote the perturbation of token score $s_t$ induced by
the low-rank residual approximation.
The page score is the log-sum-exp of its token scores, so its first-order
perturbation is
\begin{equation*}
    \delta_p
    =
    \frac{1}{\pi_p}
    \sum_{t\in\mathcal P_p}
    p_t\Delta s_t
    +
    O(\|\Delta s\|^2).
\end{equation*}
Using
$
\Delta s_t
=
q(UF^\top-I)e_t^\top/\sqrt{d_h},
$
this becomes
\begin{equation*}
    \delta_p
    =
    \frac{1}{\sqrt{d_h}}
    q(UF^\top-I)
    \left(
        \frac{1}{\pi_p}
        \sum_{t\in\mathcal P_p}
        p_t e_t
    \right)^\top
    +
    O(\|\Delta s\|^2).
\end{equation*}
We therefore define the attention-weighted mean residual within page $p$ as
\begin{equation*}
    \bar e_p
    =
    \frac{1}{\pi_p}
    \sum_{t\in\mathcal P_p}
    p_t e_t,
\end{equation*}
so that
\begin{equation*}
    \delta_p
    =
    \frac{1}{\sqrt{d_h}}
    q(UF^\top-I)\bar e_p^\top
    +
    O(\|\Delta s\|^2).
\end{equation*}

After perturbing the page scores by $\delta$, the resulting page
distribution is
\begin{equation*}
    \widetilde\pi_p
    =
    \frac{\pi_p\exp(\delta_p)}
         {\sum_j\pi_j\exp(\delta_j)}.
\end{equation*}
The KL divergence from the exact page distribution can
therefore be written as
\begin{align*}
    D_{\mathrm{KL}}(\pi\|\widetilde\pi)
    &=
    \log
    \left(
        \sum_p \pi_p\exp(\delta_p)
    \right)
    -
    \sum_p\pi_p\delta_p \\
    &=
    \frac{1}{2}
    \left[
        \sum_p\pi_p\delta_p^2
        -
        \left(
            \sum_p\pi_p\delta_p
        \right)^2
    \right]
    +
    O(\|\delta\|^3).
\end{align*}

Let
$
\bar e=\sum_p\pi_p\bar e_p
$
denote the mean residual over pages.
Substituting the first-order expression for $\delta_p$ into the quadratic
term gives
\begin{equation*}
    \frac{1}{2d_h}
    q(UF^\top-I)
    \left[
        \sum_p
        \pi_p\bar e_p^\top\bar e_p
        -
        \bar e^\top\bar e
    \right]
    (UF^\top-I)^\top
    q^\top.
\end{equation*}
The matrix in brackets is the covariance of the page residuals under the
page-attention distribution.
We denote it by
\begin{equation*}
    \Gamma
    =
    \sum_p
    \pi_p
    (\bar e_p-\bar e)^\top
    (\bar e_p-\bar e).
\end{equation*}
Dropping higher-order terms gives the local quadratic objective
\begin{equation*}
    \frac{1}{2d_h}
    q(UF^\top-I)
    \Gamma
    (UF^\top-I)^\top
    q^\top.
\end{equation*}
Restoring the calibration-query and head indices and summing over them
gives
\begin{equation}
    \mathcal L_{\mathrm{PF}}
    =
    \frac{1}{2d_h}
    \sum_{a,h}
    q_{a,h}
    (U_hF_h^\top-I)
    \Gamma_{a,h}
    (U_hF_h^\top-I)^\top
    q_{a,h}^\top.
    \label{eq:app_page_fisher_objective}
\end{equation}

\subsubsection{Alternating Updates}
\label{app:residual_solver}

For fixed residual-side factors, differentiating
\eqref{eq:app_page_fisher_objective} with respect to $U_h$ gives
\begin{equation*}
    \sum_a
    \left(
        q_{a,h}^\top q_{a,h}
    \right)
    U_h
    \left(
        F_h^\top
        \Gamma_{a,h}
        F_h
    \right)
    =
    \sum_a
    q_{a,h}^\top q_{a,h}
    \Gamma_{a,h}
    F_h.
\end{equation*}
For fixed query-side factors, differentiating with respect to $F_h$ gives
\begin{equation*}
    \sum_a
    \Gamma_{a,h}
    F_h
    \left(
        U_h^\top
        q_{a,h}^\top q_{a,h}
        U_h
    \right)
    =
    \sum_a
    \Gamma_{a,h}
    q_{a,h}^\top q_{a,h}
    U_h.
\end{equation*}
Both updates are positive-semidefinite linear systems.
We solve them by matrix-free preconditioned conjugate gradient and
alternate between the two factor families.

\subsection{Query Position Selection}
\label{app:query_selection}
For query head $h$, let $H_h=\mathbb E[q_h^\top q_h]$ be the
uncentered query second moment over the calibration set. Define
\begin{equation*}
    \widetilde q_{w,i,h}=q_{w,i,h}H_h^{\dagger/2},
    \qquad
    \phi_i=\frac{1}{\sqrt{WH_q}}
    \operatorname{concat}_{w,h}\widetilde q_{w,i,h}.
\end{equation*}
Stacking these whitened features by row gives $\Phi$ and the position
Gram matrix $G=\Phi\Phi^\top$.
We apply diagonal-pivoted Cholesky to $G$ independently within positional
strata, selecting 64 representative query positions for Page-Fisher
calibration. In exact arithmetic, this selection is equivalent to
column-pivoted QR on the corresponding whitened feature matrix
$\Phi^\top$ within each stratum.

\subsection{Runtime and Memory Accounting}
\label{app:runtime}

After the routing-aligned basis transformation in
Appendix~\ref{app:key_prediction}, the first $r_b$ coordinates of the
compressed Value contain all information required by the Value-to-Key
predictor.
Thus, no separate per-token representation is needed for the base
prediction.
The persistent routing state consists of the $r_V$-dimensional compressed
Value and the $r_q$-dimensional residual code, while the original Key is
retained separately for exact QK computation on selected pages.

\begin{table}[t]
\centering
\small
\begin{tabular}{lccc}
\toprule
State
& Persistent width
& Full-context routing
& Selected attention \\
\midrule
Compressed Value
& $r_V$
& $r_b$
& $r_V$ \\
Residual code
& $r_q$
& $r_q$
& -- \\
Exact Key
& $d_h$
& --
& $d_h$ \\
\bottomrule
\end{tabular}
\caption{Per-token state and logical read width for one physical KV head.}
\label{tab:app_runtime_cache}
\end{table}

Let $T$ be the context length and $S$ the number of tokens retained after
page selection.
If $b_V$, $b_R$, and $b_K$ denote the corresponding element sizes, the
context-dependent cache traffic per layer is
\begin{equation}
    B_{\mathrm{decode}}
    =
    H_{kv}
    \left[
        T
        \left(
            r_b b_V+r_q b_R
        \right)
        +
        S
        \left(
            d_h b_K+r_V b_V
        \right)
    \right].
    \label{eq:app_runtime_total_bytes}
\end{equation}
The first term is the full-context routing scan.
The second is paid only for selected pages, where the original Keys are
retrieved to recompute exact QK scores and the resulting attention weights
are applied to the full compressed Value states.

Without routing-aligned Value coordinates, the first term would require
reading all $r_V$ compressed-Value coordinates rather than only $r_b$.
Thus the basis transformation reduces the context-wide Value read from
$r_V$ to $r_b$ without changing the Value approximation.
Exact Keys may remain outside GPU memory because unselected Key pages are
never accessed.

\section{Experimental Details and Ablations}
\label{app:additional_experiments}
\subsection{Experimental Setup}
\label{app:exp_setup}

\paragraph{Models and Calibration.}
We evaluate Llama-3.1-8B-Instruct, Qwen3-8B-Base, Qwen3-32B,
Qwen3-8B, and Qwen3.5-9B; Qwen3-8B uses static YaRN with factor~4
for 128K routing. Value compression uses 256 C4 sequences of 2,048 tokens,
activation-aware initialization, FP32 fitting, and BF16 factors.
Llama-3.1 and Qwen3 use six ALS sweeps, 16 conjugate-gradient iterations
per encoder update, covariance damping $10^{-7}$, and joint decoder
refits. The 75\% and 50\% Value-retention settings use global rank
allocation (Section~\ref{sec:rank_allocation}). Qwen3.5-9B uses 12 sweeps
and 25\% compression of Gated-DeltaNet output projections.
Routing uses V96 for Llama-3.1-8B and Qwen3-8B, and V192 on
Qwen3.5-9B full-attention layers.

Routing calibration uses 16 C4 and 16 synthetic retrieval 128K fitting
windows, with 16 held-out C4 diagnostic windows. Each C4 window concatenates
32 document-disjoint 4,096-token excerpts. Synthetic windows insert
multi-key, multi-value, variable-tracking, and aggregation records into C4
across eight seeded positional strata, followed by a 9--13K-token segment
of 48--96 deterministic retrieval queries; identifiers are absent from the
haystack and templates are independent of the evaluation benchmarks.
Page-Fisher fitting uses 64 representative query positions
(Appendix~\ref{app:query_selection}), 40 block-coordinate sweeps,
and preconditioned conjugate gradients with at most 100 iterations and
relative damping $10^{-5}$.
Unless stated otherwise, B16R16 denotes a 16-dimensional Value-derived
Base and a 16-dimensional Key-derived residual, with Page4 routing and
64 recent tokens excluded from the routed prefix and retained exactly.

\paragraph{Evaluation Protocols.}
The main RULER-128K evaluation reports 10 tasks with 100 prompts per task,
generated by the official implementation with seed~42 and a maximum length
of 131,072 tokens. LongBench uses nine tasks: \texttt{narrativeqa},
\texttt{multifieldqa\_en}, \texttt{hotpotqa}, \texttt{musique},
\texttt{dureader}, \texttt{gov\_report}, \texttt{samsum},
\texttt{passage\_retrieval\_en}, and \texttt{lcc}.
Rendered prompts must exceed 4,096 tokens under the evaluated tokenizer,
leaving 1,543/1,549/1,564 prompts for Llama-3.1-8B/Qwen3-8B/Qwen3.5-9B.
BasisKV selects 2,048 routed tokens per KV group on RULER and 256 on
LongBench, plus 64 recent tokens. Sparse baselines share the compressed-Value checkpoint of Full-K and use the original selected Keys for exact QK computation after routing.
LRQK and Loki perform routing independently per query head. To
approximately match BasisKV's 2,048-token group-level support, we set
their per-head top-$k$ values to 832 and 856, respectively. After
deduplicating selections across query heads within each physical KV
group, the average support is 2,111.68 tokens for LRQK and 1,985.09
tokens for Loki, closely matching BasisKV's 2,048-token budget.
ShadowKV retains its native chunk, outlier, and local-support
configuration. Controlled ablations specify their exceptions below.
These top-$k$ values are chosen based on empirical support size rather
than downstream task accuracy.

\paragraph{System Measurements.}
TP8 experiments use eight 48\,GB NVIDIA L40S GPUs connected by
PCIe~4.0$\times$16 without NVLink, with NCCL through vLLM's PyNccl backend;
serving uses 4K-token prompts and 128 generated tokens at the reported
batch sizes. Sparse decode measurements use one L40S at TP1/B1 and report
medians over three trials of steady-state attention-block and full-model
decoding, excluding prefill.

\subsection{Collective-Aware Compression Ablations}
Unless stated otherwise, compression ablations use Qwen3-8B-Base with
dense Keys and dense attention.

\paragraph{Calibration robustness.}
\label{app:section3-calibration-corpus}
Refitting BasisKV and matched PaLU baselines on the same WikiText-2
training windows preserves the ordering of the compressed methods at
both ranks (Table~\ref{tab:section3-wt2-calibration}); BasisKV remains
close to dense quality at 25\% compression.

\begin{table}[!htbp]
\centering
\scriptsize
\caption{\textbf{WikiText-2 calibration robustness.}
PPL and 7-task MCQ at matched Value rank; bold marks the best compressed
method per rank.}
\label{tab:section3-wt2-calibration}
\renewcommand{\arraystretch}{1.08}
\setlength{\tabcolsep}{1.1pt}
\begin{tabular}{@{}clccccccccc@{}}
\toprule
& & & \multicolumn{8}{c}{General MCQ $\uparrow$} \\
\cmidrule(l){4-11}
Budget & Method & PPL $\downarrow$ & BoolQ & ARC-E & ARC-C & WinoG. & PIQA & HellaS. & OB & Avg. \\
\midrule
\multicolumn{2}{l}{Dense}
& 7.0025 & 0.831 & 0.802 & 0.569 & 0.728 & 0.794 & 0.786 & 0.420 & 0.704 \\
\midrule
\multirow{3}{*}{75\%}
& PaLU M & 9.0649 & 0.752 & 0.682 & 0.480 & 0.704 & 0.766 & 0.755 & 0.394 & 0.647 \\
& PaLU G4 & 9.0087 & 0.790 & 0.745 & 0.533 & 0.708 & 0.787 & \textbf{0.793} & 0.414 & 0.681 \\
& BasisKV & \textbf{7.1181} & \textbf{0.821} & \textbf{0.792} & \textbf{0.543}
& \textbf{0.734} & \textbf{0.790} & 0.766 & \textbf{0.422} & \textbf{0.695} \\
\midrule
\multirow{3}{*}{50\%}
& PaLU M & 10.9782 & 0.481 & 0.537 & 0.378 & 0.665 & 0.718 & 0.631 & 0.388 & 0.543 \\
& PaLU G4 & 9.9710 & 0.690 & 0.596 & 0.443 & \textbf{0.714} & 0.750 & \textbf{0.743} & \textbf{0.428} & 0.623 \\
& BasisKV & \textbf{7.6590} & \textbf{0.791} & \textbf{0.706} & \textbf{0.474}
& 0.694 & \textbf{0.761} & 0.706 & 0.396 & \textbf{0.647} \\
\bottomrule
\end{tabular}
\end{table}

With compression fixed within each model, changing the calibration domain
affects structured-generation quality on Qwen3.5-9B and Qwen3-32B, with
no single corpus preferred across all tasks
(Table~\ref{tab:section3-generation-calibration}).
For both models, Mixed consists of 128 C4, 64 mathematics,
and 64 code calibration windows, whereas Reasoning consists of
128 mathematics and 128 code windows.
Each window contains 2,048 tokens.
Mathematics and code prompts are drawn from OpenThoughts3,
with source datasets \texttt{ai2-adapt-dev/openmath-2-math}
and \texttt{nvidia/OpenCodeReasoning}, respectively;
responses are generated by the corresponding target model
rather than taken from the dataset's teacher answers.

\begin{table}[!htbp]
\centering
\small
\caption{\textbf{Calibration-domain transfer.}
Full-test-set task scores (\%); bold marks the best calibration corpus within
each model. Dense rows are reference scores, not calibration variants.}
\label{tab:section3-generation-calibration}
\renewcommand{\arraystretch}{1.08}
\setlength{\tabcolsep}{7.5pt}
\begin{tabular}{@{}llcccc@{}}
\toprule
& & \multicolumn{4}{c}{Accuracy (\%) $\uparrow$} \\
\cmidrule(l){3-6}
Model & Calibration & GSM8K & MATH500 & MBPP+ & IFEval \\
\midrule
\multirow{4}{*}{Qwen3.5-9B}
 & Dense & 93.25 & 91.00 & 72.75 & 83.92 \\
\cmidrule(lr){2-6}
 & C4        & 86.81          & 75.60          & 51.59          & 69.32 \\
 & Mixed     & \textbf{89.23} & 83.00          & 69.84          & \textbf{77.82} \\
 & Reasoning & 88.93          & \textbf{85.60} & \textbf{72.49} & 72.64 \\
\midrule
\multirow{4}{*}{Qwen3-32B}
 & Dense & $89.08$ & 79.60 & 45.77 & 29.21 \\
\cmidrule(lr){2-6}
 & C4        & 89.01          & 58.00          & 35.45          & 23.66 \\
 & Mixed     & 90.67          & 80.40          & \textbf{42.59} & 24.77 \\
 & Reasoning & \textbf{91.36} & \textbf{83.00} & 40.74          & \textbf{26.99} \\
\bottomrule
\end{tabular}
\end{table}

\begin{figure}[!htbp]
    \centering
    \includegraphics[width=\linewidth]{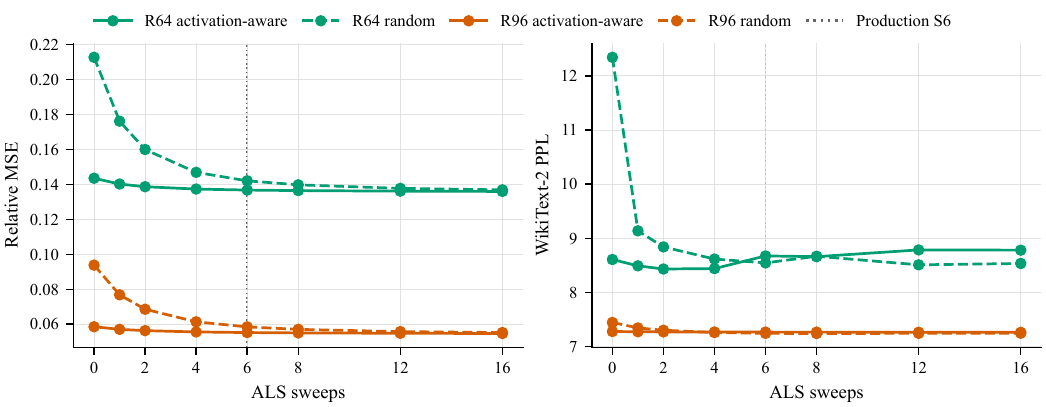}
    \caption{\textbf{ALS initialization and convergence.}
    Relative MSE (left) and WikiText-2 perplexity (right) at R64 and R96.
    The dotted line marks six ALS sweeps.}
    \label{fig:section3-als-sweeps}
\end{figure}

\paragraph{ALS convergence.}
\label{app:section3-als-ablation}
Activation-aware initialization converges faster than random orthogonal
initialization (Figure~\ref{fig:section3-als-sweeps}). Six sweeps are the
fixed production budget, with a final decoder refit for every evaluated
checkpoint; PPL is not strictly monotonic with additional sweeps.

\FloatBarrier

\paragraph{Joint fitting.}
\label{app:section3-joint-objective}
We compare BasisKV with Group-SVD, which fits each physical KV group independently using an activation-aware decomposition.
BasisKV instead fits the low-rank factors against the complete
attention output.
As shown in Table~\ref{tab:section3-joint-vs-group-svd}, the joint fit improves
held-out reconstruction error, perplexity, and average MCQ accuracy
at both ranks, although some individual tasks favor Group-SVD.

\begin{table}[!htbp]
\centering
\scriptsize
\caption{\textbf{Group-SVD versus BasisKV.}
WikiText-2 PPL and seven-task MCQ at R64/R96; bold marks the better method
per rank and metric.}
\label{tab:section3-joint-vs-group-svd}
\renewcommand{\arraystretch}{1.08}
\setlength{\tabcolsep}{1.4pt}
\begin{tabular}{@{}clcccccccccc@{}}
\toprule
& & & & \multicolumn{8}{c}{General MCQ $\uparrow$} \\
\cmidrule(l){5-12}
Rank & Method & Rel. MSE $\downarrow$ & PPL $\downarrow$
& BoolQ & ARC-E & ARC-C & WinoG. & PIQA & HellaS. & OB & Avg. \\
\midrule
\multirow{2}{*}{R64}
  & Group-SVD   & 0.16806 & 9.2547 & \textbf{0.801} & 0.717 & 0.464
                & \textbf{0.728} & 0.777 & 0.710 & 0.388 & 0.655 \\
  & BasisKV S6 & \textbf{0.13681} & \textbf{8.6747} & 0.799
                & \textbf{0.747} & \textbf{0.493} & 0.718
                & \textbf{0.783} & \textbf{0.727} & \textbf{0.400}
                & \textbf{0.667} \\
\midrule
\multirow{2}{*}{R96}
  & Group-SVD   & 0.06932 & 7.3783 & \textbf{0.806} & 0.808 & 0.529
                & \textbf{0.724} & 0.791 & 0.759 & \textbf{0.416} & 0.690 \\
  & BasisKV S6 & \textbf{0.05523} & \textbf{7.2679} & 0.799
                & \textbf{0.811} & \textbf{0.569} & 0.721
                & \textbf{0.793} & \textbf{0.768} & 0.412
                & \textbf{0.696} \\
\bottomrule
\end{tabular}
\end{table}

\FloatBarrier

\subsection{Key Routing Ablations}
\label{app:section4-ablations}
Routing ablations use Llama-3.1-8B-Instruct with V96 unless stated
otherwise; the page-granularity comparison also includes Qwen3-8B.
Calibration and default support follow Appendix~\ref{app:exp_setup}.

\paragraph{Routing-calibration robustness.}
The synthetic routing-calibration windows are constructed independently
of the evaluation benchmarks. Each 128K window consists of C4 text with
synthetic records inserted across eight positional strata, followed by
retrieval questions with deterministic answers. Record and query
templates are written independently of RULER and LongBench, and all
identifiers and values are freshly sampled. The generator does not import or reuse benchmark-specific prompts,
templates, or evaluation code. The 16
synthetic windows consist of 8 multi-key, 4 multi-value, 2
variable-tracking, and 2 aggregation windows.

For example, a multi-key window may contain records such as
\begin{quote}
\small
\raggedright
\texttt{The registry entry for ABXZ-4821 is QM-7A2F91.}\\
\emph{[C4 text and unrelated records]}\\
\texttt{Identifier RTKP-0936 has authorization code ND-C840B2.}\\
\emph{[C4 text and unrelated records]}\\
\texttt{Record: asset LWFS-7150 maps to token HK-19DE63.}\\
\emph{[additional C4 text]}\\
\texttt{Question: Retrieve the codes associated with RTKP-0936,
ABXZ-4821 and LWFS-7150, in that order.}\\
\texttt{Answer: The codes are ND-C840B2, QM-7A2F91, HK-19DE63.}
\end{quote}
The answers are included in the calibration sequence because these
windows are used only for forward-pass statistic collection rather than
as evaluation prompts.
As an additional robustness check, Table~\ref{tab:c4_only_routing}
reports an earlier B16R16 router fitted using only C4 windows.
Despite receiving no synthetic retrieval calibration, BasisKV remains
competitive with the sparse routing baselines on the full 11-task
RULER-128K suite.
This experiment uses Page32 and a different routing configuration
from the Page4 setup in the main results, so we treat it as a
robustness check rather than a controlled calibration-domain ablation.

\begin{table*}[t]
\centering
\small
\caption{
\textbf{C4-only routing robustness on RULER-128K.}
Llama-3.1-8B-Instruct with uniform V96; all methods are evaluated on
11 tasks with 100 prompts per task.
Full-K uses the same compressed V96 checkpoint.
BasisKV B16R16 is fitted using 32 purely C4 128K windows.
}
\label{tab:c4_only_routing}
\setlength{\tabcolsep}{3pt}
\renewcommand{\arraystretch}{1.12}
\resizebox{\textwidth}{!}{%
\begin{tabular}{@{}l*{11}{r}!{\vrule width 0.45pt}r@{}}
\toprule
Method & S1 & S2 & S3 & MK1 & MK2 & MQ & MV & VT & FWE & QA1 & QA2 & Avg. \\
\midrule
Full-K
& 100.00 & 99.00 & 99.00 & 99.00 & 91.00 & 97.75
& 89.75 & 66.80 & 65.67 & 76.00 & 45.00 & 84.45 \\
\midrule
Loki
& 80.00 & 94.00 & 0.00 & 92.00 & 41.00 & 55.75
& 56.50 & 20.40 & 40.67 & 53.00 & 26.00 & 50.85 \\
ShadowKV
& 96.00 & 99.00 & 96.00 & 98.00 & 84.00 & 93.25
& 87.50 & 26.60 & 54.33 & 78.00 & 45.00 & 77.97 \\
LRQK
& 100.00 & 99.00 & 99.00 & 99.00 & 88.00 & 97.50
& 91.25 & 44.40 & 44.33 & 77.00 & 45.00 & 80.41 \\
BasisKV B16R16
& 100.00 & 100.00 & 99.00 & 99.00 & 75.00 & 97.00
& 88.50 & 56.80 & 56.33 & 76.00 & 42.00 & 80.88 \\
\bottomrule
\end{tabular}%
}
\end{table*}

\begin{figure}[!htbp]
    \centering
    \includegraphics[width=0.98\linewidth]{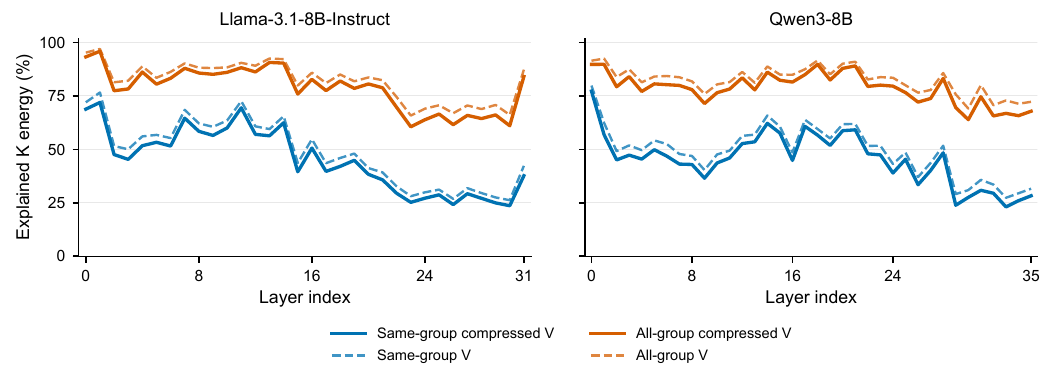}
    \caption{\textbf{Key predictability across layers.}
    Held-out centered explained pre-RoPE Key energy on Llama-3.1-8B-Instruct and Qwen3-8B.}
    \label{fig:section4-v-to-k-groups}
\end{figure}

\paragraph{Value--Key predictability.}
\label{app:v-to-k-predictability}
Affine predictors fitted on the calibration windows and evaluated held-out
measure centered explained pre-RoPE Key energy, with targets centered by
the held-out mean. These diagnostics use local or concatenated all-group
Values and are not restricted to the router's rank-16 Base.
Local compressed Values retain substantial predictive Key information,
and compressed-Value curves remain close to raw-Value curves across depth
(Figure~\ref{fig:section4-v-to-k-groups}). All-group Values provide a
higher diagnostic reference; routing uses only local Values.

\FloatBarrier

\paragraph{Base and residual components.}
\label{app:section4-components}
Table~\ref{tab:section4-base-residual} is a controlled 11-task RULER-128K
ablation with 30 shared prompts per task, separate from the main 10-task
evaluation. Base alone is insufficient, whereas R16 provides substantial
routing quality. Adding the Value-derived Base improves over R16, with a
paired 95\% bootstrap interval of $[0.80,5.17]$ points, and approaches R32
while using only 16 Key-derived residual dimensions.

\begin{table}[!htbp]
\centering
\small
\caption{\textbf{Base and Residual contributions on RULER-128K.} Llama-3.1-8B-Instruct with compressed V96. All methods use the same 30 prompts per task. Sparse routers use Page4, 2048 routed tokens and 64 recent tokens.}
\label{tab:section4-base-residual}
\setlength{\tabcolsep}{3pt}
\renewcommand{\arraystretch}{1.12}
\resizebox{\linewidth}{!}{%
\begin{tabular}{@{}l*{12}{r}@{}}
\toprule
Router & S1 & S2 & S3 & MK1 & MK2 & MQ & MV & VT & FWE & QA1 & QA2 & Avg. \\
\midrule
\rowcolor{gray!10} Full-K & 100.00 & 100.00 & 100.00 & 96.67 & 86.67 & 98.33 & 94.17 & 78.00 & 57.78 & 60.00 & 36.67 & 82.57 \\
\midrule
B16 & 70.00 & 86.67 & 16.67 & 86.67 & 23.33 & 60.00 & 25.00 & 27.33 & 32.22 & 53.33 & 33.33 & 46.78 \\
R16 & 100.00 & 100.00 & 96.67 & 96.67 & 50.00 & 97.50 & 95.83 & 64.67 & 43.33 & 60.00 & 36.67 & 76.48 \\
R32 & 100.00 & 100.00 & 96.67 & 96.67 & 73.33 & 99.17 & 94.17 & 68.00 & 50.00 & 60.00 & 36.67 & 79.52 \\
\rowcolor{blue!7} B16R16 & 100.00 & 100.00 & 96.67 & 96.67 & 76.67 & 99.17 & 95.83 & 66.00 & 43.33 & 63.33 & 36.67 & 79.48 \\
\bottomrule
\end{tabular}
}
\end{table}

\FloatBarrier

\paragraph{Page granularity.}
\label{app:page-granularity}
We compare Page1/4/8/32 on both models
(Table~\ref{tab:section4-page-granularity}); Page1 is token-level routing,
the limiting case of Page-Fisher. Finer pages favor MK2, while coarser
pages can favor FWE. Page4 has the strongest LongBench point estimate on
Llama, with paired differences from Page1 and Page8 not statistically
resolved; Qwen's highest mean is at Page1. We adopt Page4 as a balanced
operating point across task types rather than optimizing for a single
retrieval pattern.

\begin{table}[!htbp]
\centering
\scriptsize
\caption{\textbf{Page granularity across models and tasks.}
Both models use V96 and B16R16 routing.
LongBench reports nine tasks and their unweighted mean over 1,543/1,549
samples for Llama/Qwen; RULER-128K reports five common tasks with 100
shared prompts per task. Higher scores are better.
Shading marks the adopted Page4 configuration.}
\label{tab:section4-page-granularity}
\setlength{\tabcolsep}{2.6pt}
\renewcommand{\arraystretch}{1.12}
\resizebox{\linewidth}{!}{%
\begin{tabular}{@{}ll!{\vrule width 0.45pt}*{9}{r}!{\vrule width 0.45pt}r!{\vrule width 0.45pt}*{5}{r}@{}}
\toprule
& & \multicolumn{10}{c!{\vrule width 0.45pt}}{\textbf{LongBench} $\uparrow$}
& \multicolumn{5}{c}{\textbf{RULER-128K} $\uparrow$} \\
\cmidrule(lr){3-12}\cmidrule(l){13-17}
Model & Page size & NQA & MQA & HQA & MQue & DRead & GRep & SAM & PRetr & LCC & Avg.
& MK2 & FWE & VT & QA1 & QA2 \\
\midrule
\rowcolor{gray!10}
 & Full-K & 28.06 & 50.25 & 52.23 & 31.32 & 32.34 & 32.76 & 44.50 & 99.50 & 52.22 & 47.02 & 83.00 & 52.67 & 76.80 & 74.00 & 48.00 \\
\cmidrule(lr){2-17}
 & 1 & 28.79 & 50.15 & 52.18 & 29.93 & 30.77 & 32.18 & 45.23 & 99.50 & 51.24 & 46.66 & 80.00 & 46.33 & 66.00 & 75.00 & 48.00 \\
\rowcolor{blue!7}
 & 4 & 29.05 & 51.02 & 52.51 & 30.70 & 28.95 & 31.47 & 44.90 & 99.50 & 54.05 & 46.91 & 78.00 & 46.67 & 64.80 & 75.00 & 47.00 \\
 & 8 & 27.96 & 50.24 & 51.45 & 31.15 & 30.00 & 31.29 & 44.43 & 99.50 & 53.46 & 46.61 & 75.00 & 50.67 & 63.40 & 75.00 & 47.00 \\
\multirow{-5}{*}{\shortstack[l]{Llama-3.1\\8B-Instruct}} & 32 & 29.16 & 51.62 & 51.66 & 31.15 & 29.58 & 29.10 & 43.34 & 99.50 & 48.95 & 46.01 & 66.00 & 60.33 & 66.80 & 74.00 & 46.00 \\
\midrule
\rowcolor{gray!10}
 & Full-K & 25.35 & 51.47 & 51.61 & 27.34 & 17.89 & 33.53 & 46.52 & 100.00 & 62.06 & 46.20 & 61.00 & 88.00 & 93.40 & 55.00 & 36.00 \\
\cmidrule(lr){2-17}
 & 1 & 25.40 & 50.78 & 51.73 & 26.56 & 16.72 & 32.74 & 45.44 & 100.00 & 59.44 & 45.42 & 57.00 & 79.00 & 93.20 & 55.00 & 34.00 \\
\rowcolor{blue!7}
 & 4 & 26.97 & 50.94 & 51.47 & 26.63 & 16.68 & 32.86 & 44.60 & 98.50 & 58.80 & 45.27 & 53.00 & 80.33 & 93.00 & 54.00 & 33.00 \\
 & 8 & 25.47 & 50.47 & 51.28 & 25.82 & 16.03 & 32.88 & 44.63 & 98.00 & 59.64 & 44.91 & 53.00 & 82.67 & 92.40 & 52.00 & 31.00 \\
\multirow{-5}{*}{\shortstack[l]{Qwen3-8B}} & 32 & 25.51 & 48.22 & 50.80 & 25.14 & 15.03 & 31.19 & 44.16 & 88.50 & 59.80 & 43.15 & 44.00 & 88.67 & 92.00 & 49.00 & 31.00 \\
\bottomrule
\end{tabular}%
}
\end{table}

\FloatBarrier

\paragraph{Dense-Value RULER-128K.}
\label{app:section4-ruler128k-dense-v}
To isolate routing quality from Value compression, we repeat the
RULER-128K comparison with the original dense Values and output projection.
Table~\ref{tab:section4-ruler128k-dense-v} compares B16R16, ShadowKV,
LRQK, and Loki with dense attention using their evaluated sparse
configurations; support budgets are not strictly matched.
BasisKV remains close to dense and competitive with the sparse baselines,
with task-dependent differences: it is stronger on variable tracking and
frequent-word extraction than ShadowKV and LRQK, while LRQK is stronger
on multi-key, multi-query, and multi-value retrieval.

\begin{table}[!htbp]
\centering
\small
\caption{\textbf{RULER-128K with dense Values.}
Llama-3.1-8B-Instruct, original V128 and $W_O$; 100 shared prompts per task.
Scores are percentages; the mean weights all 11 tasks equally.
Bold marks the best sparse score; ties involving BasisKV highlight only BasisKV.}
\label{tab:section4-ruler128k-dense-v}
\setlength{\tabcolsep}{3pt}
\renewcommand{\arraystretch}{1.15}
\resizebox{\linewidth}{!}{%
\begin{tabular}{@{}l*{12}{r}@{}}
\toprule
\multirow{2}{*}{Method} & \multicolumn{7}{c}{NIAH} & \multirow{2}{*}{VT} & \multirow{2}{*}{FWE} & \multirow{2}{*}{QA1} & \multirow{2}{*}{QA2} & \multirow{2}{*}{Avg.} \\
\cmidrule(lr){2-8}
 & S1 & S2 & S3 & MK1 & MK2 & MQ & MV & & & & & \\
\midrule
\rowcolor{gray!10}
Dense & 100.00 & 100.00 & 99.00 & 99.00 & 89.00 & 98.75 & 94.25 & 76.20 & 74.67 & 78.00 & 46.00 & 86.81 \\
\midrule
ShadowKV & 100.00 & 100.00 & 100.00 & 99.00 & 76.00 & 97.25 & 89.50 & 70.20 & 69.33 & 79.00 & 46.00 & 84.21 \\
LRQK & 100.00 & 100.00 & 100.00 & 99.00 & \textbf{83.00} & \textbf{98.50} & \textbf{92.25} & 64.20 & 58.00 & 80.00 & 46.00 & 83.72 \\
Loki & 84.00 & 97.00 & 0.00 & 98.00 & 31.00 & 45.25 & 48.00 & 39.00 & 33.67 & 64.00 & 34.00 & 52.17 \\
BasisKV & \textbf{100.00} & \textbf{100.00} & \textbf{100.00} & \textbf{99.00} & 80.00 & 97.75 & 88.50 & \textbf{73.00} & \textbf{69.67} & \textbf{80.00} & \textbf{47.00} & \textbf{84.99} \\
\bottomrule
\end{tabular}%
}

\end{table}

\FloatBarrier

\clearpage
\subsection{Compatibility with Quantization}
\label{app:quantization}
\begingroup
\setlength{\intextsep}{8pt}
\paragraph{Value compression.}
Applying KVQuant to Keys and compressed Value latents adds only small
PPL changes at both evaluated ranks
(Table~\ref{tab:quantization-ppl}), supporting compatibility with
Value compression.

\begin{table}[!htbp]
\centering
\small
\caption{\textbf{Value compression with KVQuant.}
4-bit NUQ4 with outlier handling for Keys and Value latents, simulated
by quantize--dequantize on each BF16 checkpoint. Ranks are average retained
Value dimensions; $\Delta$PPL is relative to BF16.}
\label{tab:quantization-ppl}
\setlength{\tabcolsep}{5pt}
\renewcommand{\arraystretch}{1.12}
\begin{tabular}{@{}llrrrr@{}}
\toprule
Model & Dataset & Avg. V rank
& BF16 PPL $\downarrow$
& \shortstack{KVQuant (4-bit)\\PPL $\downarrow$}
& $\Delta$PPL \\
\midrule
\multirow{2}{*}{Qwen3-8B-Base}
& \multirow{2}{*}{WikiText-2}
& 96 & 7.283 & 7.338 & +0.055 \\
& & 64 & 8.259 & 8.277 & +0.019 \\
\midrule
\multirow{2}{*}{\shortstack[l]{Llama-3.1-8B\\Instruct}}
& \multirow{2}{*}{C4}
& 96 & 10.662 & 10.727 & +0.064 \\
& & 64 & 12.454 & 12.595 & +0.141 \\
\bottomrule
\end{tabular}
\end{table}

\paragraph{Sparse routing.}
BasisKV's LongBench average stays close to BF16 with FP8 and decreases
slightly with KVQuant (Table~\ref{tab:quantization-longbench}).
Routing remains effective when the Base is computed from dequantized
Value latents; residual codes remain BF16. Both tables use
quantize--dequantize quality simulations.

\begin{table}[!htbp]
\centering
\scriptsize
\caption{\textbf{Sparse routing with quantized KV caches.}
Llama-3.1-8B-Instruct, uniform V96, 1,543 shared LongBench samples.
Full-K: full attention; BasisKV: B16R16 Page4.
FP8: E4M3; KVQuant: NUQ4 with outlier handling.
Avg.: unweighted nine-task mean; higher is better.}
\label{tab:quantization-longbench}
\setlength{\tabcolsep}{3pt}
\renewcommand{\arraystretch}{1.15}
\resizebox{\linewidth}{!}{%
\begin{tabular}{@{}ll!{\vrule width 0.45pt}*{9}{r}!{\vrule width 0.45pt}r@{}}
\toprule
Method & KV format
& NQA & MQA & HQA & MQue & DRead & GRep & SAM & PRetr & LCC & Avg. \\
\midrule
\rowcolor{gray!10}
\multirow{3}{*}{Full-K}
& BF16
& 28.06 & 50.25 & 52.23 & 31.32 & 32.34 & 32.76 & 44.50 & 99.50 & 52.22 & 47.02 \\
& FP8
& 28.73 & 50.37 & 52.32 & 31.29 & 33.06 & 33.06 & 45.05 & 99.00 & 51.34 & 47.14 \\
& KVQuant (4-bit)
& 28.62 & 50.05 & 52.78 & 31.49 & 31.87 & 32.63 & 45.37 & 99.50 & 52.83 & 47.24 \\
\midrule
\rowcolor{gray!10}
\multirow{3}{*}{BasisKV}
& BF16
& 29.05 & 51.02 & 52.51 & 30.70 & 28.95 & 31.47 & 44.90 & 99.50 & 54.05 & 46.91 \\
& FP8
& 28.76 & 50.98 & 52.62 & 30.71 & 30.41 & 31.45 & 44.55 & 99.50 & 52.62 & 46.85 \\
& KVQuant (4-bit)
& 28.14 & 49.79 & 52.81 & 30.52 & 28.83 & 31.48 & 43.96 & 99.50 & 53.43 & 46.50 \\
\bottomrule
\end{tabular}%
}
\end{table}
\endgroup

\end{document}